\documentclass[runningheads]{llncs}

\usepackage{eccv}

\usepackage{eccvabbrv}
\usepackage{enumitem}

\usepackage{graphicx}
\usepackage{booktabs}

\usepackage[accsupp]{axessibility}  
\usepackage{listings}
\usepackage{hyperref}

\usepackage{orcidlink}

\begin{document}

\title{The Persistence of Cultural Memory: Investigating Multimodal Iconicity in Diffusion Models}

\titlerunning{Investigating Multimodal Iconicity in Diffusion Models}

\author{Maria-Teresa De~Rosa~Palmini\orcidlink{0009-0005-7700-0097} \and
Eva Cetinic\orcidlink{0000-0002-5330-1259}}

\authorrunning{De Rosa Palmini and Cetinic}

\institute{University of Zurich, Zurich, Switzerland\\
\email{maria-teresa.derosa-palmini@uzh.ch}\\
\email{eva.cetinic@uzh.ch}
}

\maketitle

\begin{center}
  \includegraphics[width=0.85\textwidth]{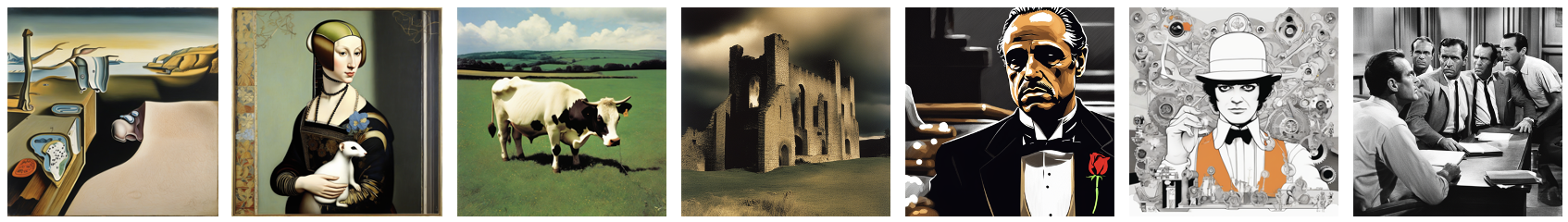}
  \captionof{figure}{Example Stable Diffusion XL generations for culturally iconic prompts: \textit{The Persistence of Memory}, \textit{Lady with an Ermine}, \textit{Atom Heart Mother}, \textit{The Unforgettable Fire}, \textit{The Godfather}, \textit{A Clockwork Orange}, and \textit{12 Angry Men}.}
  \label{fig:multimodal_examples}
\end{center}

\begin{abstract}
The ambiguity between generalization and memorization in text-to-image diffusion models becomes particularly pronounced when prompts invoke culturally shared visual references, a phenomenon we term \textit{multimodal iconicity}. These are instances in which images and texts reflect established cultural associations, such as when a title recalls a familiar artwork or film scene. Such cases challenge existing approaches to evaluating memorization, as they define a setting in which instance-level memorization and culturally grounded generalization are structurally intertwined. To address this challenge, we propose an evaluation framework to assess a model’s ability to remain culturally grounded without relying on direct visual replication. Specifically, we introduce the \textit{Cultural Reference Transformation (CRT)} metric, which separates two dimensions of model behavior: \textit{Recognition}, whether a model evokes a reference, from \textit{Realization}, how it depicts it through replication or reinterpretation. We evaluate five diffusion models on 767 Wikidata-derived cultural references, covering both still and moving imagery, and find systematic differences in how they respond to multimodal iconicity: some show weaker recognition, while others rely more heavily on replication. To assess linguistic sensitivity, we conduct prompt perturbation experiments using synonym substitutions and literal image descriptions, finding that models often reproduce iconic visual structures even when textual cues are altered. Finally, we find that cultural reference recognition correlates not only with training data frequency, but also textual uniqueness, reference popularity, and creation date. Our findings show that the behavior of diffusion models in culturally iconic settings cannot be reduced to simple reproduction, but depends on how references are recognized and realized, advancing evaluation beyond simple text–image matching toward richer contextual understanding.

\end{abstract}

\section{Introduction}
\textit{Text-to-image (TTI)} diffusion models learn complex cross-modal correspondences from massive, uncurated image–text datasets. While this scale has led to unprecedented generative capabilities, it also introduces challenges, including bias~\cite{luccioni2023stable, bianchi2023easily, palmini2025synthetic}, cultural stereotyping~\cite{ventura2025navigating, senthilkumar2024beyond}, privacy risks, and copyright violations~\cite{luccioni2023stable, somepalli2023diffusion, carlini2023extracting}. Although recent efforts have begun addressing these issues through methods such as data attribution~\cite{wang2023evaluating, brokman2024montrage} or machine unlearning~\cite{gandikota2023erasing}, a key aspect remains underexplored: distinguishing generalization from memorization when dealing with shared cultural knowledge. In specific contexts, TTI models are expected to demonstrate not only generalized world knowledge but also cultural understanding of visual and textual references. Crucially, such knowledge may intersect with copyrighted material, raising concerns about what a model should remember or forget. Therefore, balancing legitimate cultural encoding and impermissible memorization remains a fundamental challenge for~generative~AI.

Current evaluation practices reduce diverse image–text relationships to a single notion of similarity, overlooking how TTI models encode non-literal associations. For example, the prompt \textit{“The Dark Side of the Moon”} typically produces images that include prism and rainbow motifs associated with the 1973 album cover, rather than a literal lunar scene, even without explicit reference to Pink Floyd. As shown in \cref{fig:multimodal_examples}, similar patterns emerge when prompts take the form of titles of well-known paintings, album covers, or films. These examples illustrate what we term \textit{multimodal iconicity}, the culturally grounded association between words and visual motifs.

These examples highlight a setting in which instance-level memorization and culturally informed generalization are difficult to disentangle. Most existing work on memorization focuses on limiting a model’s ability to replicate training images. However, culturally iconic references introduce a different challenge: there are scenarios in which we expect the model to at least partially reproduce salient aspects of a representation, while still avoiding regurgitation due to copyright and ethical concerns. Existing evaluation metrics \cite{pizzi2022self, wang2024image} collapse these possibilities into a single notion of replication, making them poorly suited to assess how models engage with cultural memory. What is missing is a more nuanced evaluation of whether and how such associations are recognized and rendered. 

To address this gap, we introduce an evaluation framework for prompts invoking shared cultural references. Our approach disentangles two complementary aspects: whether a generation makes the intended cultural reference recognizable, and how that reference is visually realized. This distinction allows us to separate culturally grounded reinterpretation from closer forms of visual reuse. We formalize this perspective in the \textit{Cultural Reference Transformation (CRT)} metric, which captures the balance between recognizability and originality, and validate it against human judgments to ensure alignment with how people perceive cultural reference and visual reuse.

In summary, our key contributions are:
\begin{itemize}[leftmargin=*, itemsep=2pt]
    \item We introduce and formalize \emph{multimodal iconicity}, culturally grounded text–image associations, as a new evaluation dimension for TTI diffusion models. 
    
   \item We develop a evaluation framework that separates cultural reference \textit{recognition} from its \textit{realization}, distinguishing informed reinterpretation from direct replication, and validate both dimensions through human evaluation.
 
    \item We apply this framework to five diffusion models, four open-source (Stable Diffusion~2, XL, and 3~\cite{stabilityai2022sd21,podell2023sdxl,esser2024scaling},  Flux Schnell~\cite{fluxschnell}), and one proprietary (Imagen~4~\cite{imagen4}), evaluated on 767 Wikidata-derived cultural concepts~\cite{vrandevcic2014wikidata} including both \textit{still} imagery (canonical works such as paintings or album covers) and \textit{moving} imagery (multi-instance media such as films or series).
    
    \item We show that cultural reference recognition depends not only on training-data exposure, but also on factors such as \emph{textual uniqueness} and \emph{creation date}, providing new insight into how diffusion models internalize cultural references.
    
\end{itemize}

\section{Related Work}
Recent studies show that diffusion models can memorize and reproduce their training data~\cite{somepalli2023diffusion,carlini2023extracting}, raising major privacy and copyright concerns~\cite{jiang2023ai,zhang2023counterfactual,carlini2022privacy}. To mitigate these risks, several approaches for investigating and detecting replication have been developed. Data-centric methods reduce redundancy and overfitting in large-scale corpora through deduplication~\cite{carlini2023extracting,webster2023duplication} and data augmentation~\cite{daras2023ambient} techniques. Model-centric approaches instead target memorization directly, for example by randomizing captions to weaken text-conditioned replication~\cite{somepalli2023diffusion}, adjusting cross-attention or prediction magnitudes to detect and suppress memorized prompts~\cite{wen2024detecting,ren2024unveiling}, or quantifying replication into a probability density function~\cite{wang2024image}. Recent advances have further refined these evaluations by exploring the underlying mechanics of memorization, including early-stage denoising trajectories~\cite{kim2025diffusion}, localized patch retrieval~\cite{chen2024exploring}, and non-local replication triggers~\cite{kowalczuk2025finding}. Transparency-driven approaches complement these efforts by improving interpretability through data attribution, linking generated outputs to influential training samples~\cite{brokman2024montrage,wang2023evaluating}, and machine unlearning, which removes information about specific data or concepts~\cite{gandikota2023erasing,zhang2024defensive,zhang2024forget}. To standardize evaluation, recent benchmarks assess the removal of intellectual property~\cite{moon2025holistic}, while automated frameworks leverage the reasoning capabilities of VLMs to identify and mitigate copyright infringement~\cite{liu2025copyjudge,xu2025can}. Despite these advances, replication is largely treated as a technical problem to eliminate, overlooking its cultural dimensions, which we investigate through the concept of multimodal iconicity.

The cultural aspect of text-to-image relations in TTI models has been discussed by Cetinić~in \cite{cetinic2022myth}, who indicates that risk-mitigation techniques such as deduplication, while preventing "regurgitation," may also erase meaningful cultural associations. This perspective aligns with the emerging framework of computational hermeneutics, which conceptualizes generative models as cultural technologies whose outputs must be interpreted rather than simply measured \cite{kommers2025computational}, resonating with views of AI systems as models of culture~\cite{underwood2021mapping}. Yet so far, evaluations of TTI models have primarily addressed culture in relation to stereotypes, often focusing on geographically grounded depictions of people and events \cite{ventura2025navigating, malakouti2026culture, 10.1145/3613904.3642877}, while culture in the context of art has mainly been examined through style mimicry and copyright \cite{ shan2023glaze, 10.1145/3600211.3604681} rather than as a system of culturally established artifacts and references.

As a result, concepts such as iconicity remain underexplored in computational research on visual media. Among the first to introduce iconicity into computer vision, Saleh and van Noord~\cite{saleh2022computational} find that iconic photographs (e.g. \textit{Migrant Mother}) are not inherently more memorable, suggesting that iconicity is shaped by shared cultural context rather than perceptual salience. Extending the exploration of iconicity to generative AI, van Noord and Garcia~\cite{van2025iconicity} used a curated subset of iconic images to assess whether these images disproportionately influence TTI models, and find no evidence of such an effect, suggesting a difference between how human and generative AI models approach visual culture. While these existing studies focus on the iconicity of the image itself, we instead investigate iconicity as a property of the relationship between images and texts.

\section{Dataset of Iconic Image-Text Pairs}
To the best of our knowledge, no dataset exists specifically designed to empirically measure multimodal iconicity. Because a large-scale empirical study of globally recognized iconicity is beyond the scope of this work, we approximate it through indicators of cultural prominence derived from knowledge-graph metadata. More specifically, to create a dataset of iconic image–text pairs, we use Wikidata~\cite{vrandevcic2014wikidata}, a structured knowledge graph that enables querying cultural entities by semantic type (e.g., films, TV series, artworks), from which we retrieve concepts by instance type. Our selection of prompts includes only titles, omitting artist names or other explicit cues. Additionally, to focus only on word–visual relationships rather than identity cues, we remove titles that contain named entities (e.g., \textit{Mona Lisa}) using spaCy, thus minimizing the influence of highly distinctive proper nouns and reducing the lexical uniqueness signal previously shown to facilitate retrieval-like memorization \cite{somepalli2023understanding}. We then leverage \textit{Wikidata sitelinks}, cross-language links connecting a Wikidata entity to its corresponding Wikipedia articles across different languages, as a proxy for cross-cultural visibility, retaining references with more than 20 sitelinks as a data-driven indicator of global prominence across linguistic communities. The resulting dataset contains two categories of references: (1) \textit{still-image references}, each associated with a single canonical visual representation (artworks, albums, photographs), and (2) \textit{moving-image references}, associated with multiple possible visual realizations (films, TV series, animation), which share recognizable visual or thematic cues. By applying this criterion, we identified 767 cultural references (374 still and 393 moving-image ones).  Analyzing DiffusionDB \cite{wang2023diffusiondb}, we find that titles corresponding to the 767 cultural references in our dataset appear in about 6\% of prompts, indicating that culturally referential prompting is common in everyday TTI use.  Further dataset details are provided in \cref{sec:appendix_dataset} of the suppl.~material.

It is important to note that the dataset reflects Wikidata's coverage biases, which primarily document Anglophone and Western contexts. Consequently, the cultural references in our benchmark may underrepresent non-Western forms of iconicity. Addressing this will require future datasets with broader cultural coverage, though our prompt-agnostic framework readily extends to such settings.

\section{Methodology}

\label{sec:method}
To operationalize the concept of multimodal iconicity, our evaluation framework distinguishes two parts: (i) Recognition, assessing whether a generated image evokes the intended cultural reference; (ii) Realization, examining how that reference is visually realized. We formalize these dimensions into the \textit{Cultural Reference Transformation (CRT)} metric, which is designed to reward models that not only "remember" the cultural reference, but are also able to demonstrate it through new, independent imagery rather than mere replication.

\vspace{-2mm}

\begin{figure*}[ht!]
\centering
\includegraphics[width=0.90\textwidth]{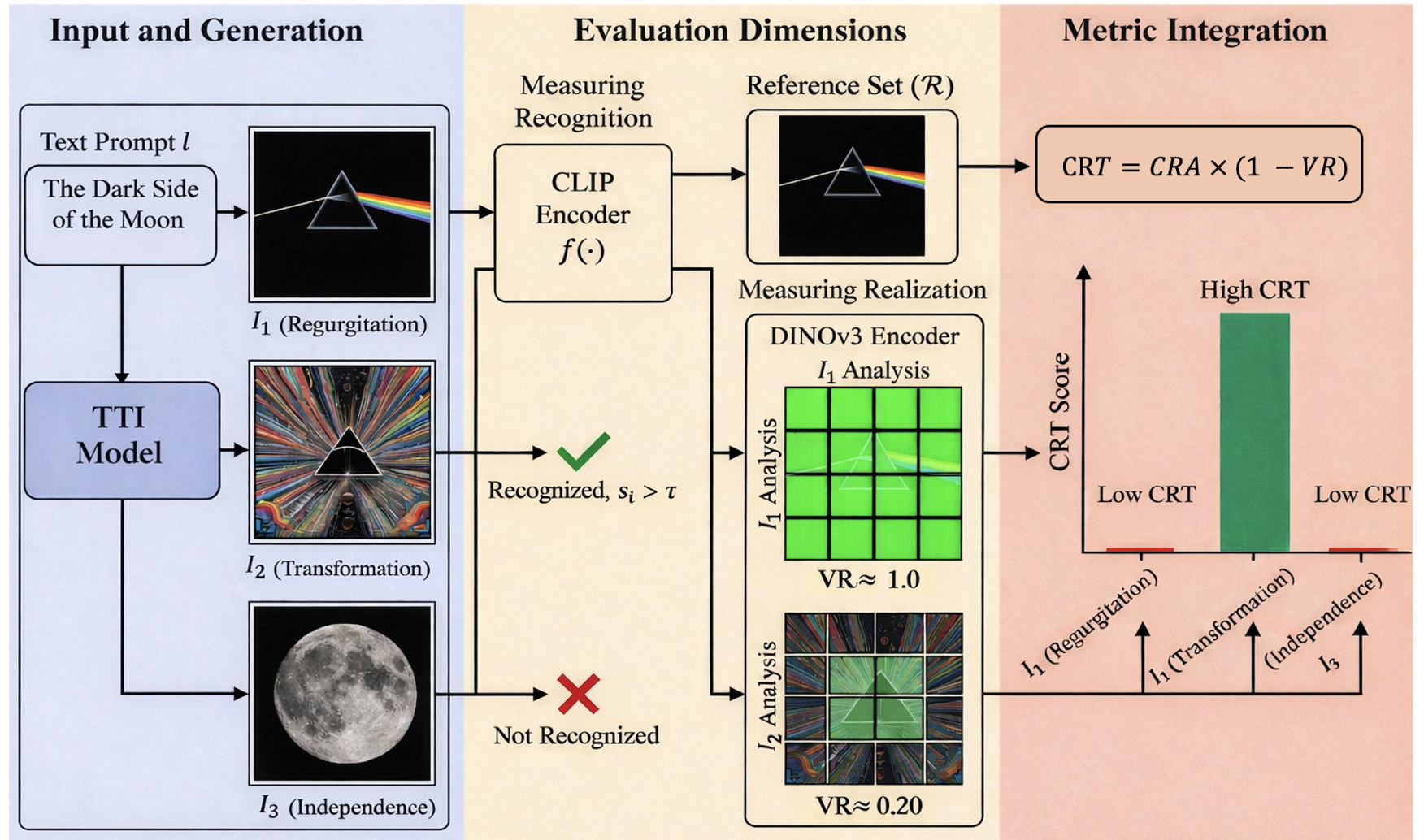}
\caption{
\textbf{Framework for evaluating multimodal iconicity.}
Generated images are evaluated along two axes: \textit{Recognition}, quantified by the CLIP-based \textit{CRA} score, and \textit{Realization}, operationalized through the patch-level \textit{VR} score. CRT integrates these axes to distinguish regurgitation ($I_1$), transformation ($I_2$), and independence ($I_3$).
}
\vspace{-3mm}
\label{fig:framework_overview}
\end{figure*}

\vspace{-6mm}

\subsection{Measuring Recognition}

To assess reference recognition, i.e., whether the generated image evokes the intended cultural reference, we compute cosine similarity between CLIP ViT-B/32   ~\cite{radford2021learning} embeddings of generated and reference images. For each cultural reference, each model generates $n=10$ images using the reference title as the prompt, with different random seeds. These outputs are compared to a reference set $\mathcal{R}$ representing canonical visual depictions of the concept. For \textit{still-image references}, $\mathcal{R}$ consists of a single canonical Wikipedia image, whereas for \textit{moving-image references}, it contains a set of representative images depicting the reference. To create this set,  we retrieve the top 50 Google Image results using the reference title as the search query and filter a visually coherent subset by removing images whose average CLIP similarity to the remaining images is below 0.7. 

For each generated image $I_i$, we compute a similarity score measuring the semantic alignment between the generation and the reference imagery. 
For still-image references, this is the cosine similarity  to the single canonical image $s_i =  \cos\!\big(f(I_i), f(R)\big)$, where $f(\cdot)$ denotes the CLIP ViT-B/32 image encoder. For moving-image references, we compute $s_i = \max_{R_j \in \mathcal{R}} \cos\!\big(f(I_i), f(R_j)\big)$, where the maximum selects the reference image most similar to the generated image. A generation is considered \textit{recognized} if $s_i > \tau$, where $\tau = 0.7$, is a recognition threshold empirically selected to balance false positives and true matches  (see \cref{sec:thresholds} of the supl. material). As illustrated in Fig.~\ref{fig:framework_overview}, generations prompted by \textit{The Dark Side of the Moon}, such as $I_2$, preserve iconic motifs (e.g., a triangular prism) and exceed the recognition threshold despite compositional variation, whereas outputs such as $I_3$, depicting a literal lunar scene, fall below the threshold and are not considered recognized. We summarize \textit{recognition} by aggregating this criterion across all generated images per cultural reference,  and refer to this quantity as the  \textit{Cultural Reference Alignment} score ($\mathrm{CRA}$):
\begin{equation}
\mathrm{CRA} = \frac{1}{n}\sum_{i=1}^{n}\mathbf{1}[s_i > \tau].
\label{eq:recognition}
\end{equation}
This ratio-based formulation is preferred over averaging similarity scores, since means can obscure variation across generated images, e.g., a few strongly recognized outputs may be diluted by many unrelated ones, yielding values that are difficult to interpret. Reporting \textit{recognition} as a proportion provides a more interpretable measure: for example, a $\mathrm{CRA}$ value of $0.6$ indicates that 60\% of generated images evoke the intended cultural reference. Additionally, we perform CRA backbone robustness tests reported in \cref{sec:cra_encoders} of the suppl. material.

\subsection{Measuring Realization}
Having identified recognized generations ($s_i > \tau$), we next assess reference realization for these outputs, capturing how the reference is visually rendered. We operationalize realization through the \textit{Visual Reuse ($\mathrm{VR}$)} score that measures the extent of localized visual content reused from the reference imagery. As illustrated in Fig.~\ref{fig:framework_overview}, this approach enables us to distinguish regurgitation ($I_1$) from partial iconic motif reuse ($I_2$), a distinction that global similarity metrics can obscure by averaging localized details into a single value.

Following \cite{somepalli2023diffusion}, we use DINOv3~\cite{simeoni2025dinov3}, which achieves state-of-the-art performance on instance retrieval benchmarks and provides finer discrimination of local correspondences. Both generated and reference images are divided into a $4{\times}4$ grid, yielding $K=16$ non-overlapping patches per image. Let $\mathcal{P} = \{p_1, \dots, p_K\}$ denote the DINOv3 patch embeddings of a generated image, and let $\mathcal{Q} = \{q_1, \dots, q_M\}$ be the pool of patch embeddings extracted from reference images in $\mathcal{R}$, where $M = K \times |\mathcal{R}|$. For still-image references $|\mathcal{R}| = 1$ and thus $M = K$, whereas for moving-image references $|\mathcal{R}| > 1$. For each generated patch $p_k$, we compute its maximum cosine similarity to the reference pool irrespective of spatial position: $s_k = \max_{q_m \in \mathcal{Q}} \cos(p_k, q_m)$. We consider the patch visually reused if $s_k > \tau_{\text{patch}}$, where $\tau_{\text{patch}} = 0.6$, is selected empirically (see \cref{sec:thresholds} of the suppl. material for further details). For a generated image, visual reuse is first measured as the proportion of its patches that satisfy this criterion. The overall $\mathrm{VR}$ score for a cultural reference is then obtained by averaging this proportion  across the $n_r$ recognized generated images: 


\begin{equation}
\mathrm{VR} = \frac{1}{n_r K}\sum_{i \in \mathcal{I}_r}\sum_{k=1}^{K} \mathbf{1}[s_k^{(i)} > \tau_{\text{patch}}]
\end{equation}

Lower values indicate independent visual synthesis, whereas higher values reflect stronger reliance on localized replication across the model’s outputs.

\subsection{Cultural Reference Transformation (CRT)} 
To integrate both recognition and realization into a single measure tailored to multimodal iconicity, we introduce the \emph{Cultural Reference Tranformation} metric. CRT captures the balance between a model’s ability to evoke a cultural reference and the extent to which it visually reuses material from canonical depictions. As illustrated in \cref{fig:framework_overview} this formulation distinguishes three characteristic outcomes: \textit{regurgitation} ($I_1$) where a reference is recognized but closely replicated; transformation ($I_2$) where the reference is recognizable yet visually reinterpreted; and \textit{independence} ($I_3$), where the reference is not recognized.  We define CRT as:

\begin{equation}
\mathrm{CRT} = \mathrm{CRA} \times (1 - \mathrm{VR}),
\label{eq}
\end{equation}
where $\mathrm{CRA}$ is the Cultural Reference Alignment score and $\mathrm{VR}$ is the Visual Reuse score. This formulation yields high values only when a model reliably evokes the intended reference while limiting direct visual reuse. It thus favors generations that preserve culturally recognizable motifs while introducing visual variation, reflecting successful transformation rather than direct replication.


\section{Results}

\subsection{Comparison with Existing Replication Metrics}
\label{sec:replication_comparison}
To evaluate how our framework relates to existing approaches for detecting replication in diffusion models, we compare against two widely used baselines or studying training-data memorization: SSCD~\cite{pizzi2022self}, which measures global embedding similarity, and PDF-Embedding (PDFE), introduced by ICDiff~\cite{wang2024image}, which predicts perceptual resemblance on a 0--5 scale. 

We first examine whether VR captures the extent of reused visual content more precisely than global similarity metrics. Using 100 still-image references, we generated synthetic image pairs under four controlled overlap conditions, exact copy, 50\% overlap, 25\% overlap, and unrelated pairs, by replacing a predefined proportion of patches in a target image with patches copied from a source image (see \cref{sec:VR_validation} of the suppl. material for further construction details). Across these conditions, VR scales closely with the true proportion of reused content (0.97, 0.51, 0.27, and 0.02, respectively), whereas SSCD and PDFE exhibit greater dispersion at intermediate overlap levels. This indicates that global replication metrics can detect 
broad resemblance, but are less reliable for quantifying how much visual material is actually reused when replication is partial or localized.

Additionally, we compare CRA, VR, and CRT across PDFE similarity levels for all models to examine how perceptual similarity relates to cultural reference alignment and transformation (see \cref{sec:pdfe_variation} in the suppl. material). At intermediate PDFE levels (2–4), both CRA and CRT span nearly the entire [0, 1] range, suggesting that similar global replication scores can arise from fundamentally different scenarios. Even at the highest replication level (PDFE = 5), VR values vary widely (0.04–0.93), indicating high perceptual similarity does not necessarily correspond to localized visual reuse. Qualitative examples in \cref{sec:pdfe_variation} illustrate how PDFE conflates replication with other forms of resemblance, while our decomposition into CRA, VR, and CRT provides a clearer framework for analyzing how TTI models balance the trade-off between visual reuse and transformation.

\subsection{Model-Level Comparison}
\label{sec:results}

\Cref{tab:static_dynamic_vr_vrk_crtbin} summarizes aggregate model performance on still and moving-image references. We report mean CRA, mean VR of recognized references, and mean CRT. We also report VR@k, which measures visual reuse on the subset of references recognized by all models, where $k$ is the number of such references ($k=110$ for still images; $k=186$ for moving images). Restricting the analysis to this shared subset removes variation in recognition coverage and enables comparison of VR across models under matched recognition conditions. Pairwise differences were assessed on matched per-reference values using two-sided Wilcoxon signed-rank tests with Holm correction for multiple comparisons. Corresponding mean paired differences and bootstrap 95\% confidence intervals are reported in \cref{sec:appendix_model_stats}. 

\begin{table}[ht!]
\label{tab:model_comparison}
\centering
\caption{\textbf{Aggregate model performance on still- and moving-image references.} Values are mean~$\pm$~SD computed across references for each model; arrows indicate direction of improvement and bold denotes the best value in each column.}

\label{tab:static_dynamic_vr_vrk_crtbin}
\scriptsize
\setlength{\tabcolsep}{4pt}
\renewcommand{\arraystretch}{1.1}

\vspace{2pt}
\textbf{(a) Still-Image References}\\[3pt]
\begin{tabular}{lcccc}
\toprule
\textbf{Model} & \textbf{CRA $\uparrow$} & \textbf{VR $\downarrow$} & \textbf{VR@k $\downarrow$} & \textbf{CRT $\uparrow$} \\
\midrule
Flux Schnell & 0.401          & \textbf{0.108 $\pm$ 0.151} & \textbf{0.108 $\pm$ 0.151} & 0.358 $\pm$ 0.023 \\
Imagen 4     & \textbf{0.623} & 0.281 $\pm$ 0.266          & 0.244 $\pm$ 0.251          & \textbf{0.448 $\pm$ 0.020} \\
SD2          & 0.489          & 0.263 $\pm$ 0.228          & 0.258 $\pm$ 0.223          & 0.361 $\pm$ 0.021 \\
SD3          & 0.535          & 0.165 $\pm$ 0.218          & 0.170 $\pm$ 0.225          & \textbf{0.447 $\pm$ 0.023} \\
SDXL         & 0.572          & 0.254 $\pm$ 0.235          & 0.239 $\pm$ 0.243          & 0.426 $\pm$ 0.022 \\
\bottomrule
\end{tabular}

\vspace{6pt}
\textbf{(b) Moving-Image References}\\[3pt]
\begin{tabular}{lcccc}
\toprule
\textbf{Model} & \textbf{CRA $\uparrow$} & \textbf{VR $\downarrow$} & \textbf{VR@k $\downarrow$} & \textbf{CRT $\uparrow$} \\
\midrule
Flux Schnell & 0.679          & 0.245 $\pm$ 0.170          & 0.251 $\pm$ 0.172          & 0.512 $\pm$ 0.019 \\
Imagen 4     & 0.816          & 0.212 $\pm$ 0.157          & 0.216 $\pm$ 0.154          & \textbf{0.643 $\pm$ 0.016} \\
SD2          & 0.867          & 0.289 $\pm$ 0.194          & 0.311 $\pm$ 0.198          & 0.617 $\pm$ 0.015 \\
SD3          & \textbf{0.875} & 0.281 $\pm$ 0.179          & 0.304 $\pm$ 0.189          & 0.629 $\pm$ 0.014 \\
SDXL         & 0.684          & \textbf{0.191 $\pm$ 0.215} & \textbf{0.187 $\pm$ 0.210} & 0.553 $\pm$ 0.021 \\
\bottomrule
\end{tabular}
\end{table}

\begin{figure}[ht!]
    \centering
\includegraphics[width=0.95\linewidth]{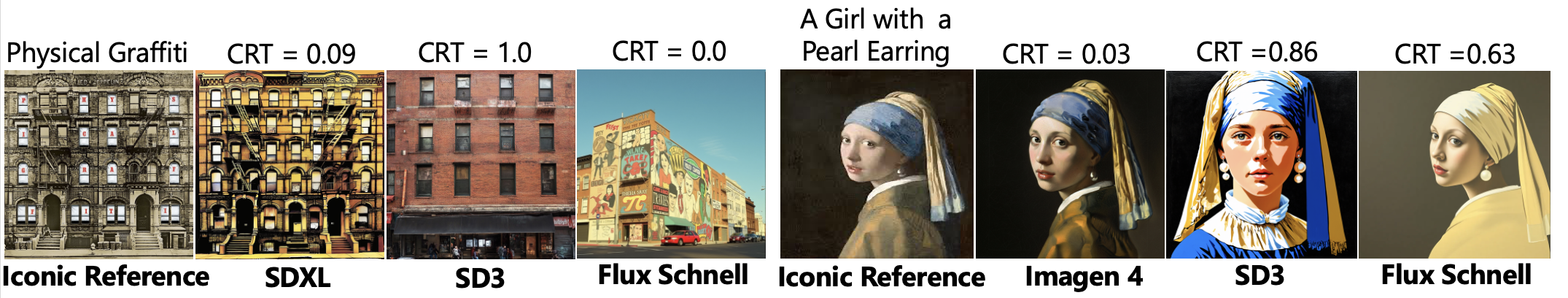}
\caption{Example of images generated from the prompts \emph{Physical Graffiti} and \emph{A Girl with a Pearl Earring} using three diffusion models for each cultural reference, \textit{SDXL}, \textit{SD3}, \textit{Flux Schnell} and \textit{Imagen 4}, shown alongside the iconic cultural reference image.}
    \label{fig:examples_pdfe}
    \vspace{-5mm}
\end{figure}

\noindent\textbf{Still-Image References.} \textit{Imagen 4} achieves the highest recognition rate (CRA) ($p < 0.0001$), followed by \textit{SDXL} and \textit{SD3}, whereas \textit{Flux Schnell} recognizes substantially fewer references. However, models with higher recognition often exhibit greater visual reuse: both \textit{Imagen 4} and \textit{SDXL} exhibit significantly higher reuse (VR and VR@k) than \textit{SD3} on jointly recognized references ($p < 0.05$). At the same time, \textit{Flux Schnell} represents the opposite extreme, showing the lowest reuse overall but also the lowest recognition. CRT integrates these contrasting behaviors. In the final scores, \textit{Imagen 4} and \textit{SD3} form a statistically indistinguishable top tier ($p > 0.61$): \textit{Imagen 4} reaches high CRT through stronger recognition, whereas \textit{SD3} achieves comparable performance through lower visual reuse, as shown in the qualitative examples of \cref{fig:examples_pdfe}. By comparison, \textit{SDXL} does not reach this tier because its reuse remains relatively high without \textit{Imagen 4}'s recognition advantage, while \textit{Flux}'s low reuse similarly does not translate into high CRT due to its substantially lower recognition.

\noindent\textbf{Moving-Image References.} \textit{SD3} and \textit{SD2} achieve the highest recognition rates (CRA $\approx 0.87$, $p < 0.0001$), but also exhibit significantly higher visual reuse (VR and VR@k) than \textit{Imagen 4} and \textit{SDXL} ($p < 0.001$). In comparison, \textit{SDXL} and \textit{Flux Schnell} exhibit lower reuse but recognize fewer references overall. \textit{Imagen 4} achieves the highest transformation score ($CRT = 0.643$), balancing strong recognition with substantially lower reuse than the leading SD models. As a result, \textit{Imagen 4} outperforms both \textit{SD2} and \textit{SD3} in the final CRT analysis ($p < 0.005$), whereas \textit{SDXL} and \textit{Flux} remain limited by their lower CRA scores.

Overall, CRT identifies which models best balance recognition and visual reuse, while CRA and VR clarify the underlying trade-offs. Additional qualitative examples illustrating these behaviors, are provided in \cref{sec:qualitative_examples}.

\subsection{Relation between Recognition and Visual Reuse}

\begin{figure*}[ht!]
    \centering
    \includegraphics[width=\textwidth]{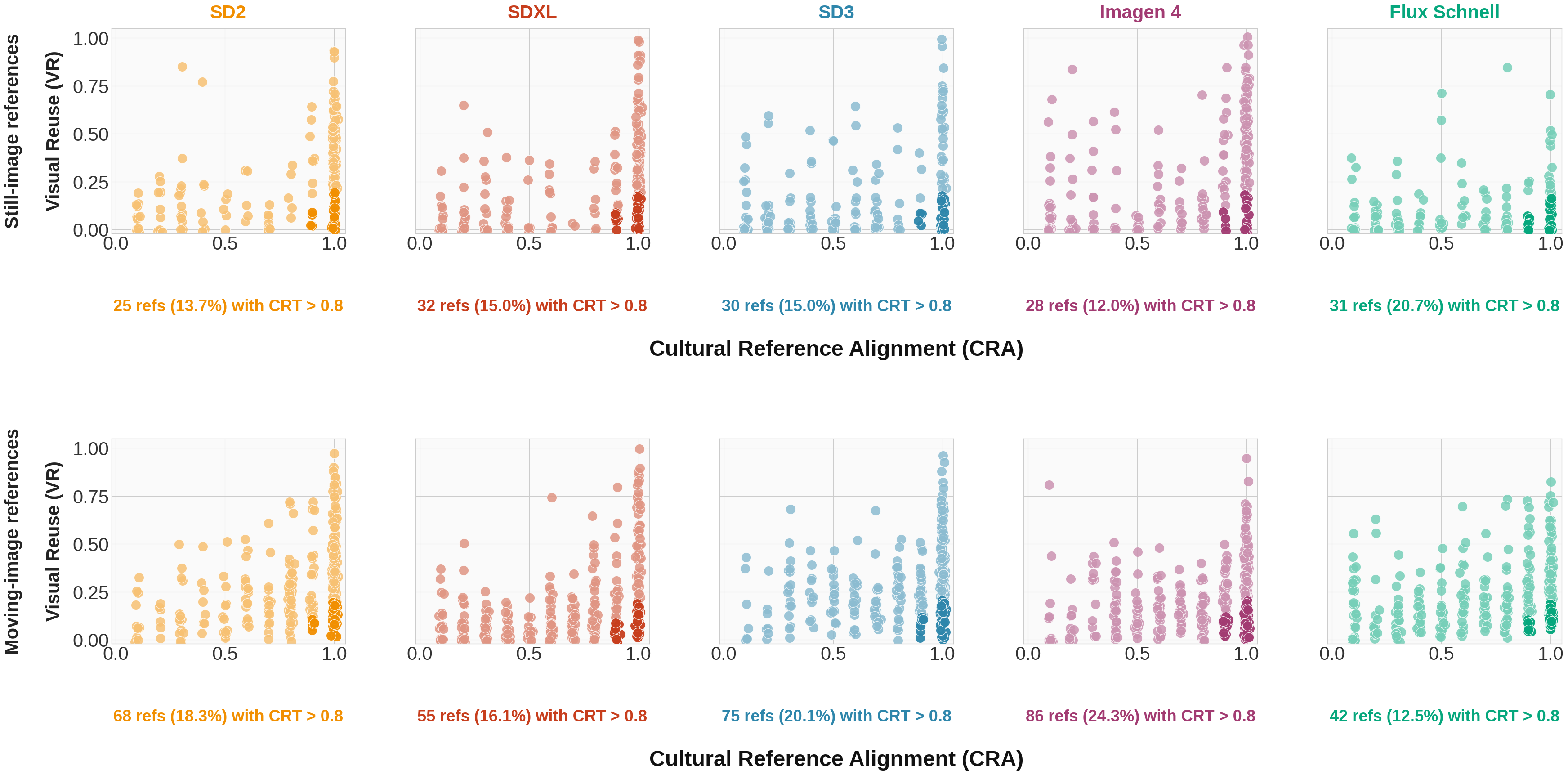}
    \caption{
        CRA--VR relationship across diffusion models for still (top) and moving (bottom) image references. Each point corresponds to a single reference, and shows its CRA and VR scores.
        Darker points indicate high CRT (\(> 0.8\)), where the reference is recognized while the generated image remains visually independent. Percentages below each subplot denote the proportion of recognized references achieving high CRT. }
        \vspace{-4mm}
    \label{fig:cra_vr_tradeoff}
\end{figure*}

To evaluate the relationship between recognition and visual reuse, \cref{fig:cra_vr_tradeoff} maps the relationship between CRA and VR, revealing that while these metrics exhibit moderate positive correlation overall (Spearman’s $\rho = 0.58$, $p < 0.001$ for still images; $\rho = 0.40$, $p < 0.001$ for moving images), visual reuse varies widely across references at high recognition levels. Among references with CRA $> 0.8$, visual reuse shows substantial dispersion  ($SD = 0.247$ for still images; $SD = 0.198$ for moving images), suggesting that conceptual generalization remains uncommon. Across all models, only 12--27\% of recognized references exhibit high CRT ($> 0.8$), where a model successfully generates the defining visual motif of a cultural reference without directly copying the original image. For still images with high recognition, \textit{Imagen 4}, \textit{SD2}, and \textit{SDXL} exhibit consistently high visual reuse  (mean $\approx 0.34$--$0.42$), whereas \textit{Flux Schnell} maintains significantly lower scores (mean = 0.15, $p < 0.0001$), suggesting greater visual transformation. Conversely, this pattern shifts for moving-image references, where \textit{Imagen 4} demonstrates the lowest visual reuse among highly recognized subset (mean = 0.24) compared to the higher reuse rates of other models (mean $\approx 0.29$--$0.32$). Ultimately, this decoupling demonstrates that cultural reference recognition can occur without direct visual replication, though the degree to which models successfully separate a concept from its original form varies substantially across different architectures.

\subsection{Human Evaluation}
\label{sec:human_evaluation}

We validate the two aspects of CRT, \textit{recognition} and \textit{realization}, against human judgments using an evaluation set of 1,000 image pairs derived from generations prompted by still-image references. We focus on still-image references; as it captures the core behaviors of recognition and visual reuse. These were randomly sampled from generated images produced by all five models and stratified to cover low, medium, and high $\mathrm{VR}$ scores (based on VR quantiles). The sample ensures that each still-image reference appears at least once and includes 80\% recognized and 20\% non-recognized candidates according to the CRA recognition criterion. Each pair was rated by three independent annotators, yielding 3,000 total annotations from 300 crowd-sourced participants. For every pair, annotators were shown a reference image alongside a generated image and asked: (Q1) whether the generated image evoked the intended cultural reference, and (Q2) the extent of visual content reuse on a five-point scale (see \cref{sec:human_evaluation} for detailed task instructions).

Our results demonstrate that both metrics align closely with human judgment: for Q1, CRA achieves high accuracy (0.91), precision (0.94), and recall (0.95) against majority-voted labels, maintaining a stable consensus signal despite moderate inter-annotator agreement (Fleiss' $\kappa = 0.44$). For Q2, $\mathrm{VR}$ correlates strongly with human ratings (Spearman’s $\rho = 0.71, p < 0.001$), with robust alignment in high-reuse ($r = 0.84$) and low-reuse ($r = 0.68$) regimes. Although the intermediate regime is more variable ($r = 0.35$) reflecting the difficulty of judging partial transformations, correlations remain significant ($p < 0.001$), supporting the framework’s ability to quantify cultural reference transformation.

\subsection{Effects of Textual Variation  }
\label{sec:prompt_perturbations}

To determine whether cultural reference recognition is tied to specific titles or grounded the prominence of specific visual motifs, we conducted two controlled prompt perturbation experiments: a \textit{synonym variant}, where \textit{Llama-3.2-11B} replaced exactly one content word with a semantically close substitute (e.g., \textit{"The Shriek"} for \textit{"The Scream"}), and a \textit{literal description} obtained via a Visual Question Answering (VQA) setup using \textit{Llama-3.2-Vision 11B}. These models were selected for their ability to preserve referent anonymity and semantic fidelity during the generation process  (see \cref{sec:llm_prompts} for further details).

\begin{figure*}[t!]
  \centering
  \begin{subfigure}[b]{0.48\textwidth}
    \centering
    \includegraphics[width=\linewidth]{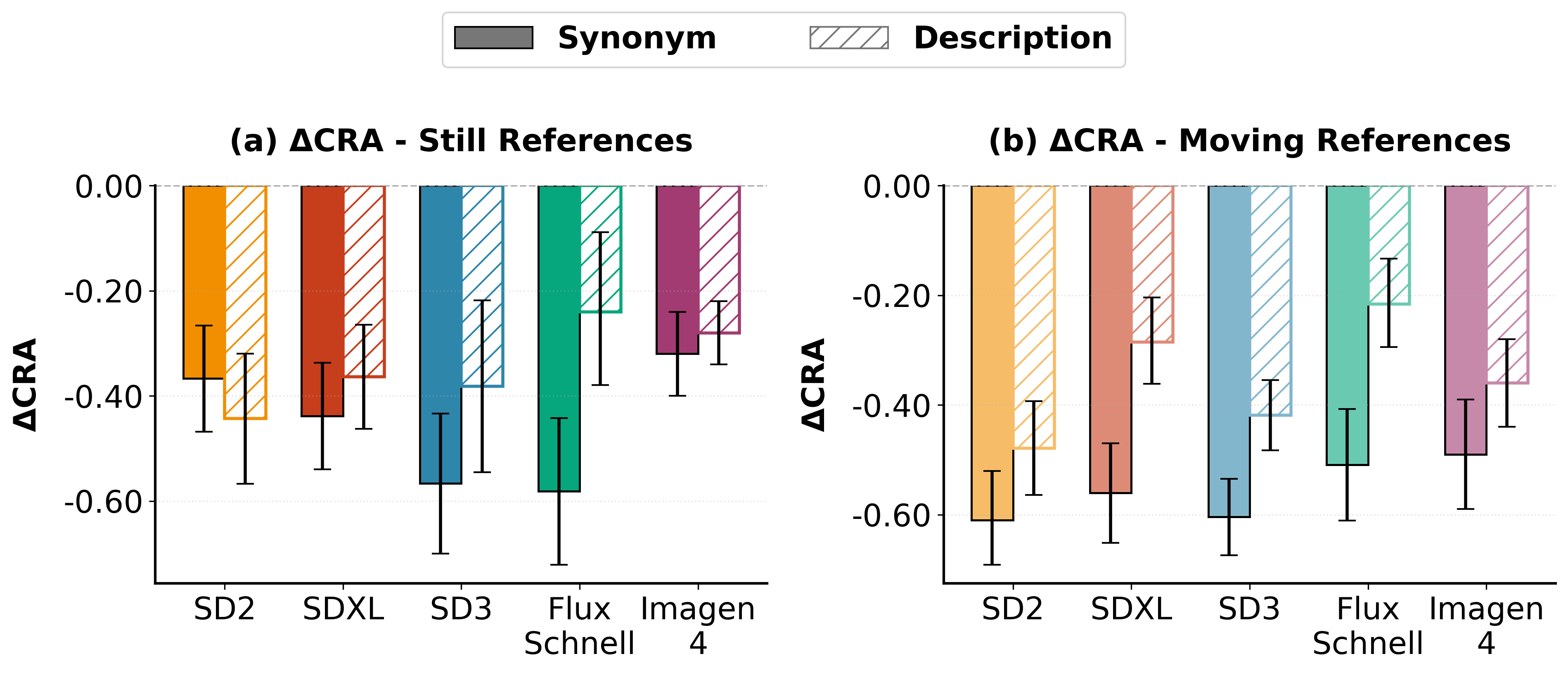}
    \caption{Change in Cultural Reference Alignment ($\Delta$CRA) under prompt perturbations.}
    \label{fig:delta_cra}
  \end{subfigure}
  \hfill
  \begin{subfigure}[b]{0.48\textwidth}
    \centering
    \includegraphics[width=\linewidth]{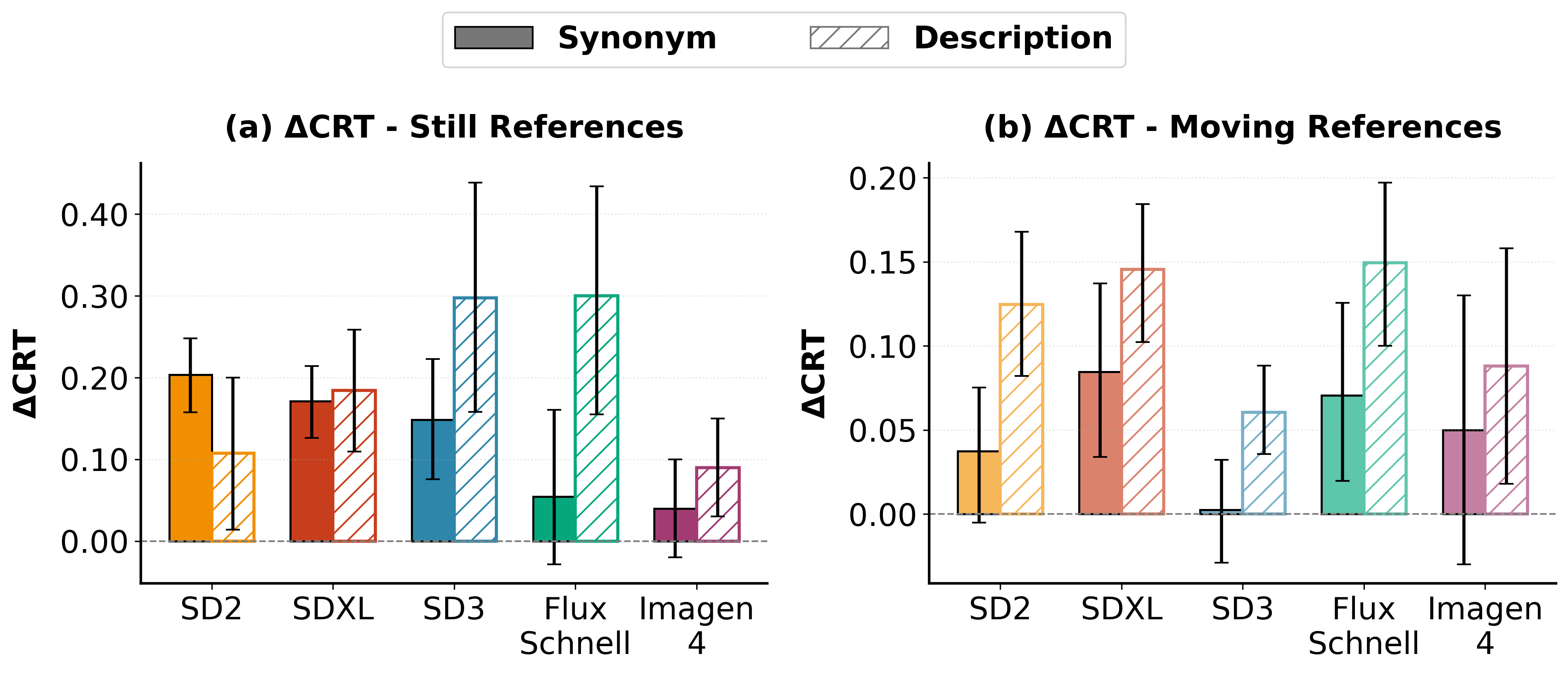}
    \caption{Change in Transformation ($\Delta$CRT)  of recognized references under prompt perturbations. }
    \label{fig:delta_crt_retained}
  \end{subfigure}

  \caption{
    \textbf{$\Delta$CRA and $\Delta$CRT under prompt perturbations.} Mean changes in \textit{recognition} and \textit{transformation} for still and moving references. $\Delta$CRT is computed only when references remain recognizable ($\text{CRA} > 0$). Error bars show 95\% CIs.
  }
\end{figure*}

\begin{figure*}[ht!]
  \centering
  \begin{subfigure}[t]{0.485\textwidth}
    \centering
    \includegraphics[width=\linewidth]{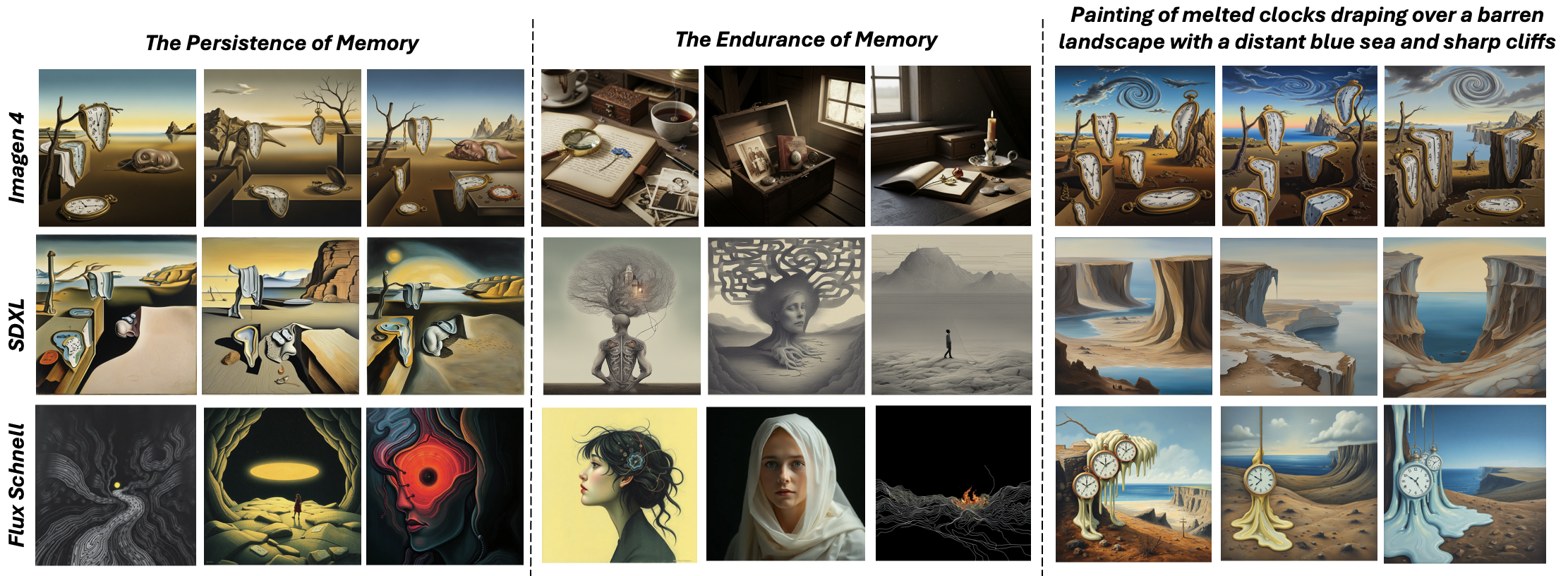}
    \caption{\textbf{The Persistence of Memory.}}
    \label{fig:persistence_memory}
  \end{subfigure}\hfill
  \begin{subfigure}[t]{0.485\textwidth}
    \centering
    \includegraphics[width=\linewidth]{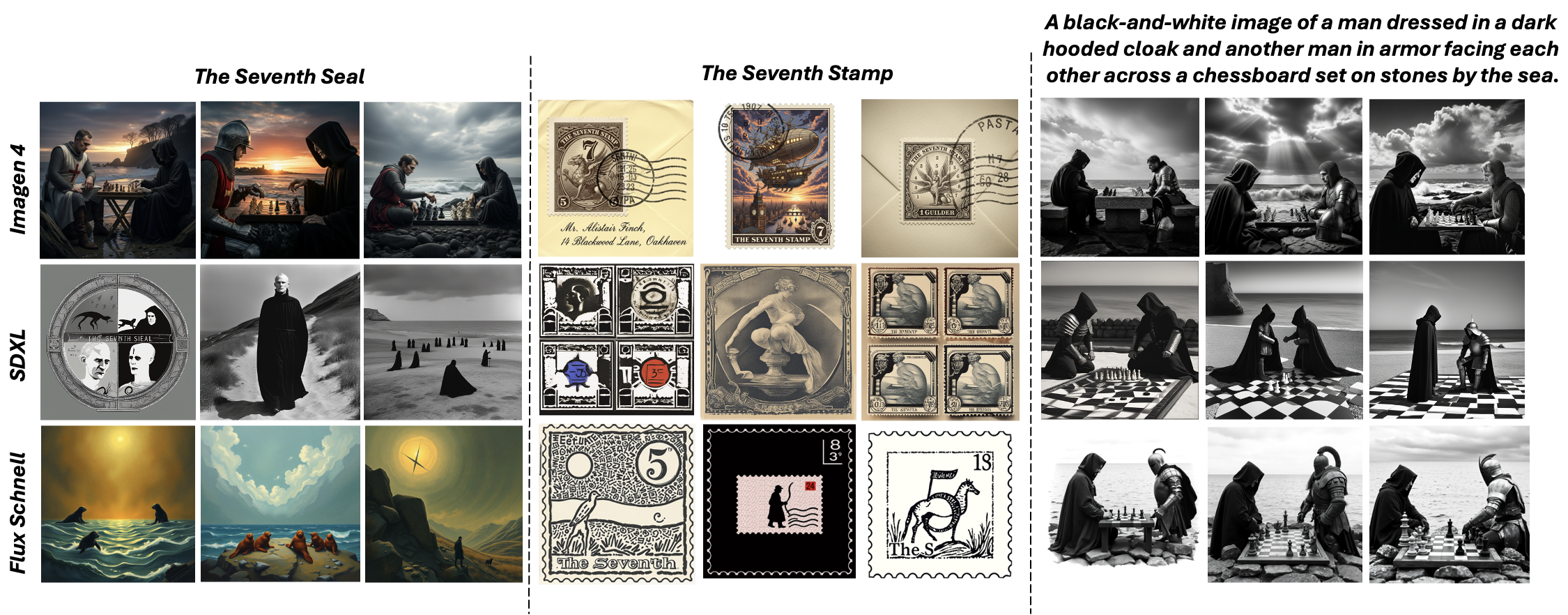}
    \caption{\textbf{The Seventh Seal.}  }
    \label{fig:seventh_seal}
  \end{subfigure}
  \vspace{-1pt}
  \caption{Prompt perturbation examples. Generations for still (\textit{The Persistence of Memory}) and moving (\textit{The Seventh Seal}) references}
  \vspace{-3mm}
  \label{fig:qualitative_perturbations}
\vspace{-1pt}
\end{figure*}

As shown in \cref{fig:delta_cra}, in both perturbation experiments, we observe consistent declines in CRA, indicating that all models become less likely to evoke the intended reference once the prompt wording is altered.. The effect is systematic yet varies in magnitude: description prompts generally produce smaller drops than synonym substitutions, suggesting that richer visual–semantic context can partially compensate for lexical variations, aligning with prior findings that richer captioning can improve TTI generation performance under varied prompt formulations \cite{esser2024scaling}. Qualitative results shown in \cref{fig:qualitative_perturbations} support this, showing that synonym prompts often cause outputs to diverge from the iconic motif, while descriptive prompts tend to preserve core visualstructures. Among the tested models, \textit{Imagen 4} retains the highest CRA  across both preturbation variants, indicating a superior capacity to reproduce iconic visuals despite altered linguistic cues. Finally, for the subset of references that remain recognized after perturbation, we observe a consistent decrease in VR alongside a significant increase in CRT ($p < 0.05$, \cref{fig:delta_crt_retained}), suggesting that recognized generations under perturbed prompts are rendered with greater visual transformation.

\vspace{-5mm}

\subsection{Factors Influencing Reference Recognition }
To understand why diffusion models succeed or fail at recognizing cultural references, we examine which factors are most strongly linked to variation in CRA. In this analysis, we focus on SD v2.1~\cite{stabilityai2022sd21}, a widely used open-source model trained on a deduplicated and filtered subset of LAION-5B~\cite{schuhmann2022laion}. Because the exact training set composition is not publicly available, we approximate the model’s training distribution using the open LAION-400M subset~\cite{schuhmann2021laion} as a coarse proxy for the model’s training data. Our goal is to understand which factors influence whether a model recognizes iconic image–text associations: (i) training-data-related features, indicating how each reference is represented in this coarse training distribution, and (ii) reference-related features, capturing traits of each example in our dataset of iconic image–text pairs.

\begin{figure*}[ht!]
\centering
\makebox[\textwidth][c]{%
  \begin{minipage}{\textwidth}
    \centering
    \begin{subfigure}[t]{0.24\textwidth}
      \includegraphics[width=\linewidth]{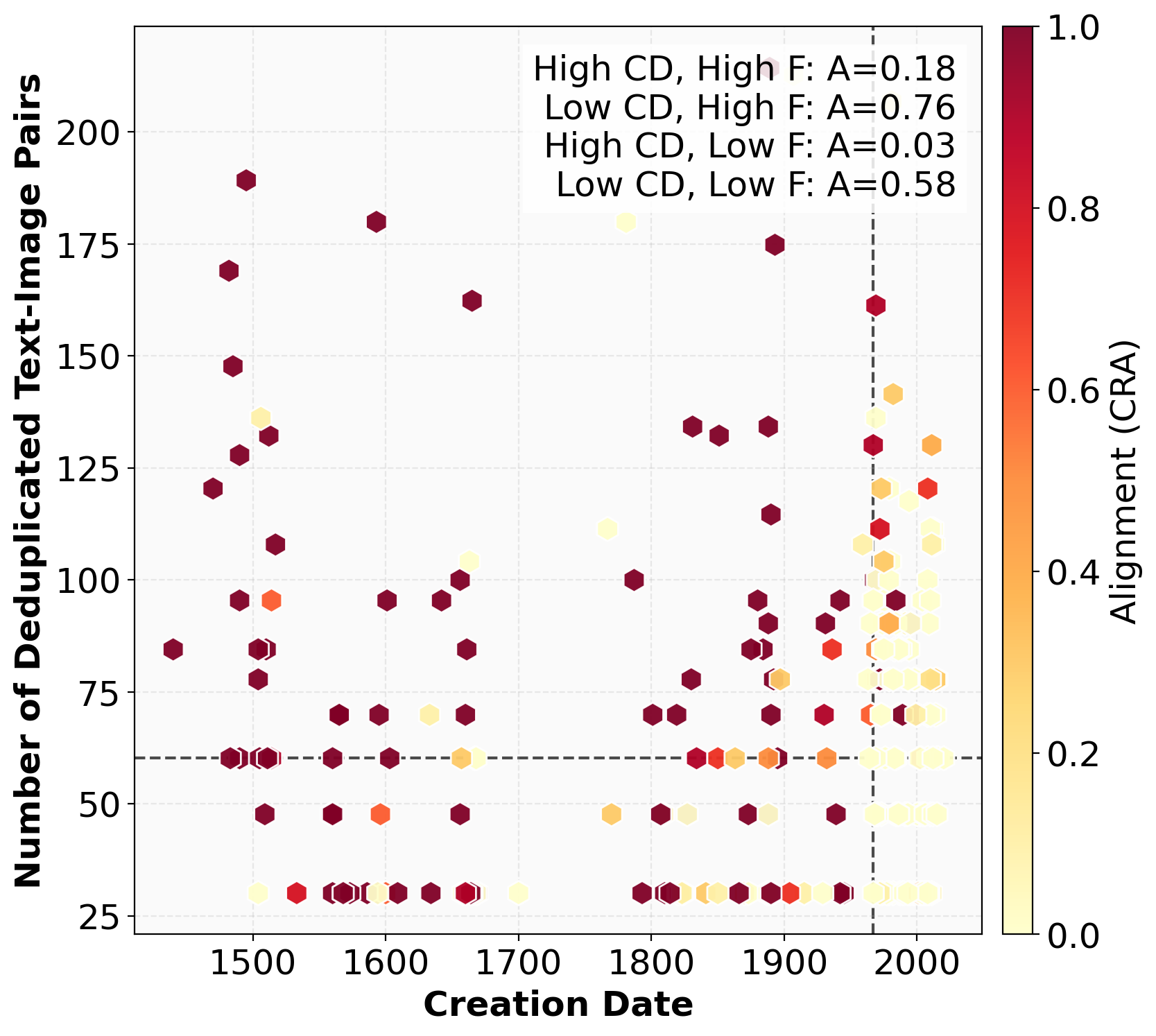}
      \caption{Creation Date (Still-images)}
    \end{subfigure}\hfill
    \begin{subfigure}[t]{0.24\textwidth}
      \includegraphics[width=\linewidth]{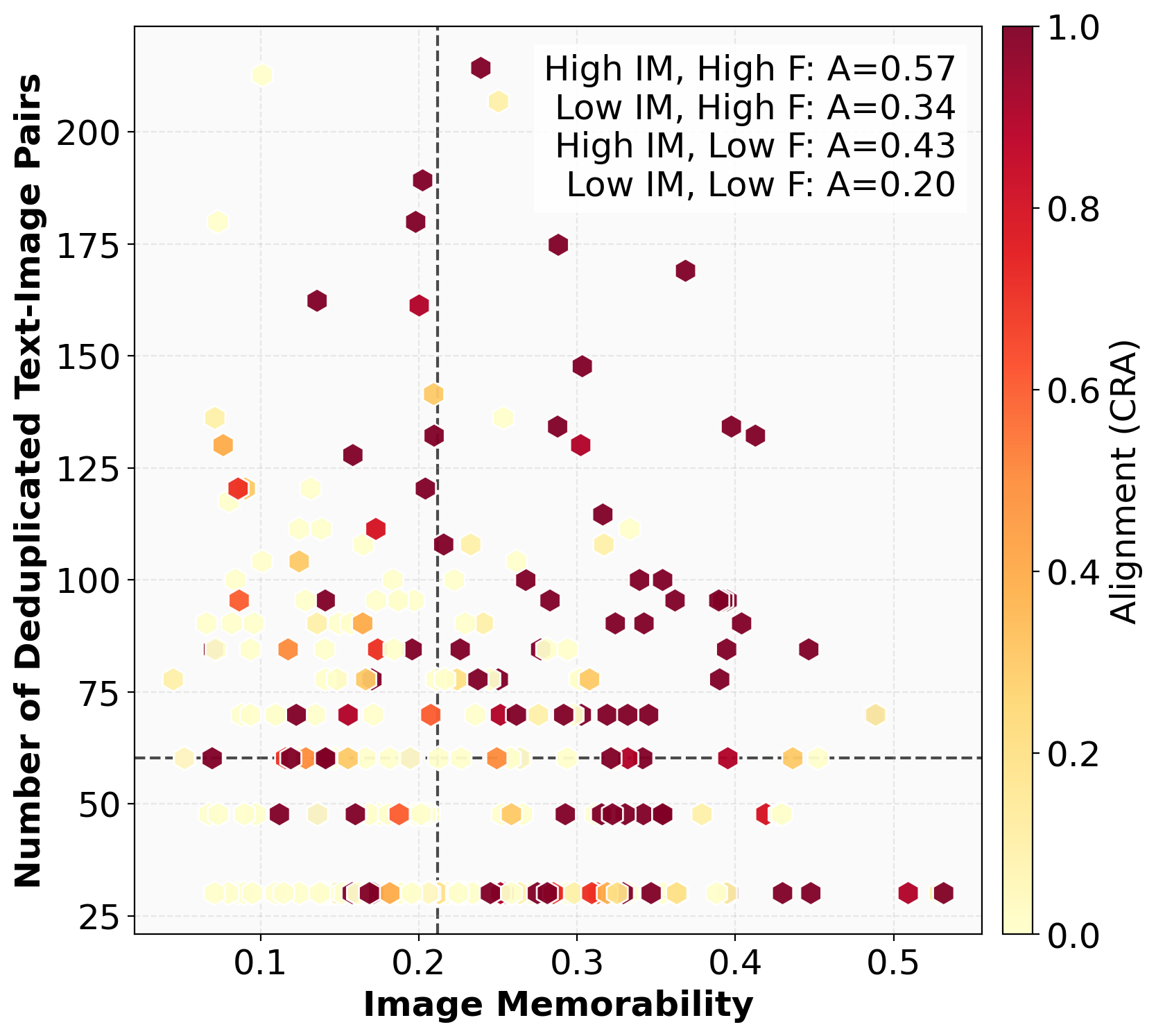}
      \caption{Image Memorability (Still-images)}
    \end{subfigure}\hfill
    \begin{subfigure}[t]{0.24\textwidth}
      \includegraphics[width=\linewidth]{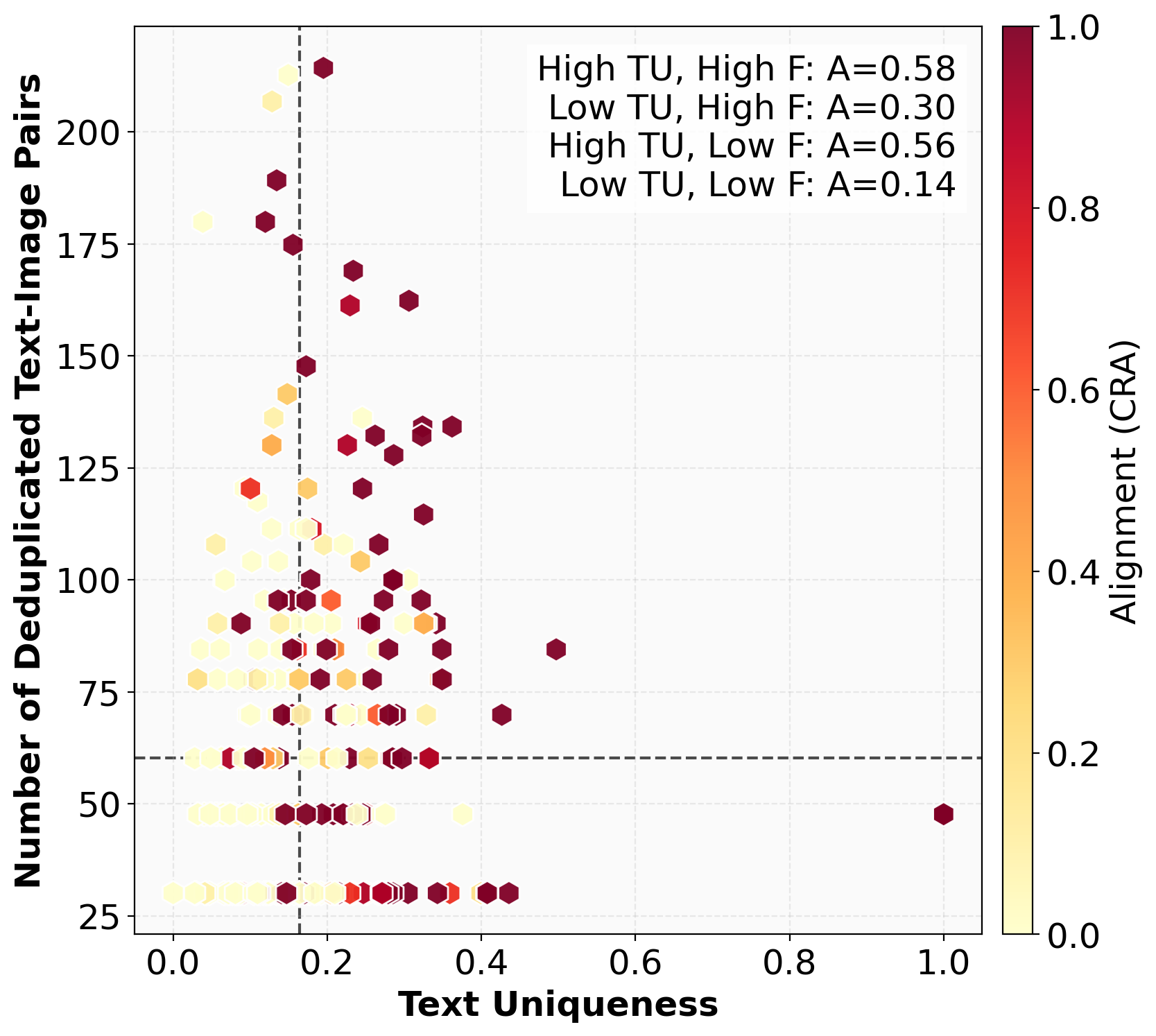}
      \caption{Text Uniqueness (Still-images)}
    \end{subfigure}\hfill
    \begin{subfigure}[t]{0.24\textwidth}
      \includegraphics[width=\linewidth]{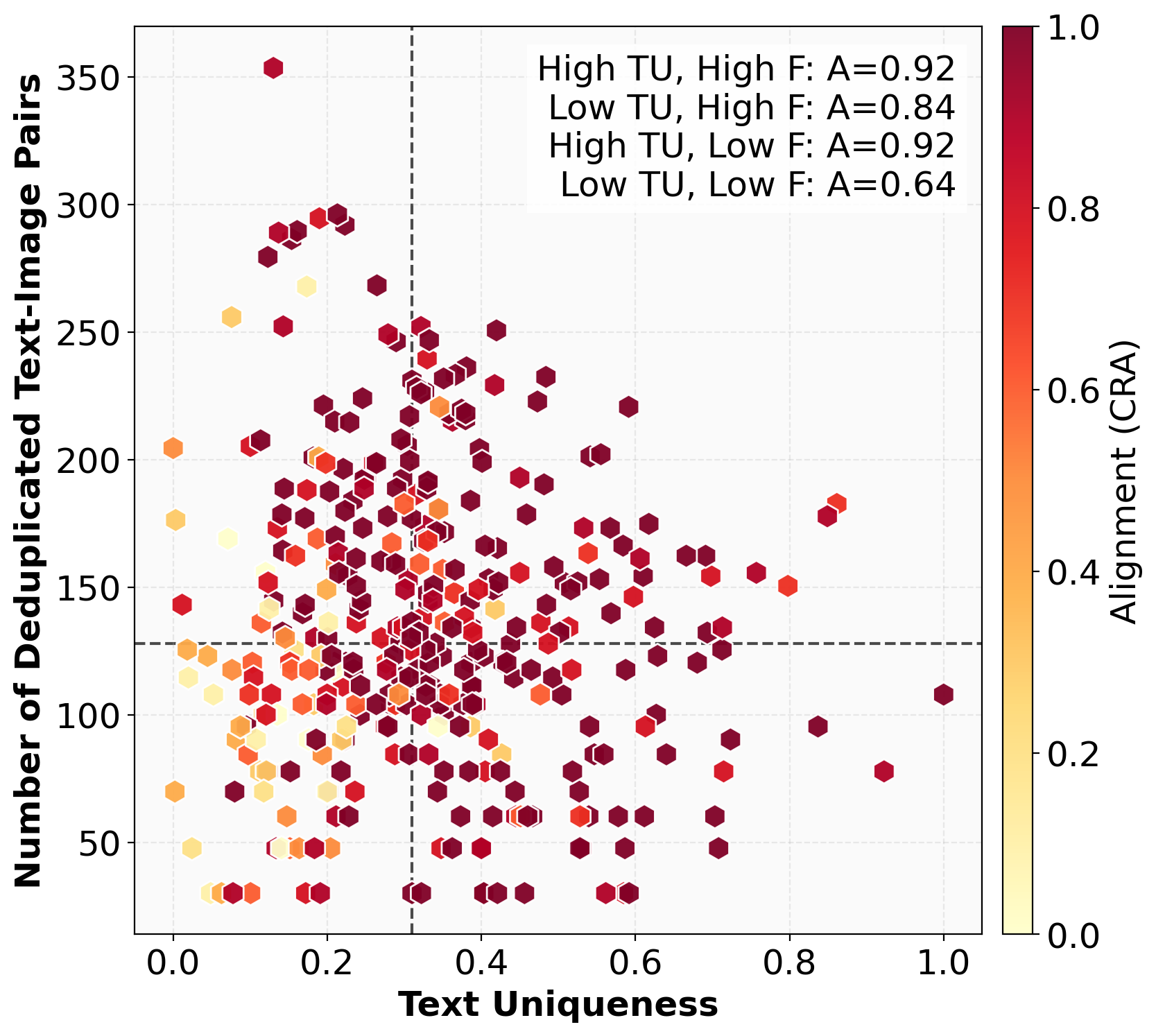}
      \caption{Text Uniqueness (Moving-images}
    \end{subfigure}
  \end{minipage}
}
\vspace{-3pt}
\caption{
\textbf{Strongest correlates of CRA as a function of the number of deduplicated text–image pairs.}
Each scatterplot shows how CRA varies with \textit{creation date}, \textit{image memorability}, and \textit{text uniqueness} in relation to the number of deduplicated text–image pairs. 
Points are colored by CRA, and median splits along both axes define quadrants annotated with average CRA values. 
}
\vspace{-3mm}
\label{fig:univariate_cra}
\end{figure*}

\paragraph{Training Data$-$Related Features}
To estimate training presence for each cultural reference, we retrieve LAION-400M samples whose captions contain the reference title and whose images are similar to the iconic depiction ($t > 0.7$). These searches return both \textbf{near-duplicates} of the reference and \textbf{related but not duplicate} content (e.g., merchandise or derivative imagery). We remove near-duplicates using SSCD ($>0.90$), following standard practice~\cite{somepalli2023diffusion,somepalli2023understanding}, and focus on the (i)~\textbf{Number of related non-duplicate images}, which more accurately reflects how the reference appears in the training data (see~\cref{sec:dedup_examples}). Additionally, to assess how distinct each reference is within the broader LAION embedding space, we use precomputed CLIP (ViT-B/32) embeddings and run a similarity search via FAISS~\cite{douze2025faiss} to derive two metrics: (ii)~\textbf{Text uniqueness}, measuring the average dissimilarity between the reference title and its nearest textual neighbors, and (iii)~\textbf{Image uniqueness}, computed analogously using the iconic reference image. Higher values indicate that a reference has fewer close textual or visual neighbors and is thus more distinctive in the embedding space.

\paragraph{Cultural Reference$-$Related Features}  We focus on five features that capture different aspects of each reference: (i)~\textbf{Popularity}, measured by the number of Wikidata sitelinks as a proxy for visibility; (ii)~\textbf{Time of release}, the year of creation or publication; (iii)~\textbf{Image memorability}, predicted using ResMem~\cite{needell2022embracing}, which estimates how intrinsically memorable the image is; (iv)~\textbf{Word memorability}, based on human recognition accuracies from memorability norms ~\cite{tuckute2025intrinsically}, and (v)~\textbf{Text concreteness}, computed as the average concreteness score of the words in the reference title using the psycholinguistic norms of~\cite{brysbaert2014concreteness}.

To identify the factors that best explain variation in CRA, we computed Spearman correlations between CRA and the features described above.  (see \cref{sec:correlation_analysis} for more details). The strongest correlate is \textit{text uniqueness}: this holds for both still-image ($\rho=0.50$, $p<0.001$) and moving-image  ($\rho=0.44$, $p<0.001$) references, aligning with recent findings that caption specificity can act as a retrieval cue for memorized training examples in diffusion models ~\cite{somepalli2023understanding}. \textit{Text concreteness} also shows a weak positive correlation ($\rho=0.16$), suggesting a minor trend in which more abstract titles hinder recognition. Among still images, \textit{creation date} shows the highest correlation ($\rho=-0.63$), with older cultural works achieving higher CRA, potentially reflecting the stronger online presence and repeated reproduction of canonical artworks. \textit{Image memorability} ($\rho=0.32$) also correlates positively with CRA, suggesting that visually distinctive motifs are learned more reliably by the model.

\begin{figure}[t!]
    \centering
    \includegraphics[width=0.95\linewidth]{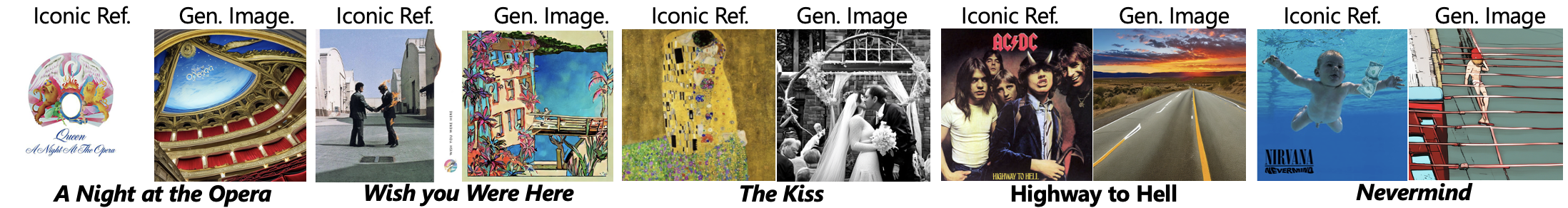}
    \caption{
        \textbf{Examples of low text uniqueness cultural references with no alignment in SD~v2.1.} Iconic references (\emph{right}) and corresponding generations (\emph{left}). 
        All shown references have text uniqueness below $0.1$ and exhibit near-zero CRA.
    }
    \vspace{-3mm}
    \label{fig:examples_pdfe}
\end{figure}

While the number of deduplicated text–image pairs correlates positively with CRA, the effect is modest. As visualized in \cref{fig:univariate_cra}, high CRA values cluster where both feature strength and training presence are high, while low values appear where both are weak . This suggests that CRA depends not just the presence of a reference in the training dat , but also on the distinctiveness of a reference’s textual and visual cues. 
For instance, examples such as \textit{A Night at the Opera}, \textit{Wish you Were Here}, and \textit{The Kiss }occur frequently in LAION yet exhibit low caption uniqueness ($<0.1$) and near-zero CRA (\cref{fig:examples_pdfe}), highlighting that data quantity alone does not guarantee cultural reference alignment.

\section{Conclusion}
We introduced a framework for evaluating how diffusion models engage with culturally iconic image–text relations, separating reference recognition from visual reuse. Our results indicate that the concept of multimodal iconicity captures a nuanced aspect of the generalization–memorization trade-off often missed by standard similarity metrics. Evaluating recognition and realization separately reveals that models align with cultural references in different ways: some rely on visual replication, while others produce transformed yet recognizable imagery. Our analysis further shows that alignment depends not only on training-data presence but also on the distinctiveness of textual cues.

Several limitations of our study should be acknowledged. Our dataset reflects Wikidata’s Western and Anglophone biases, and extending the framework to a more culturally diverse references remains an important direction of future work. Training‐data factors were analyzed only for SD2 using LAION-400M as a proxy, as the training compositions of other models are undisclosed; which makes these findings indicative rather than definitive. Moreover, our correlation analyses capture associations rather than causal mechanisms, and more controlled experiments are needed to determine how dataset properties shape cultural alignment. 
Our human evaluation focuses on still-image references; extending this validation to moving-image references would require a more complex experimental setup due to their greater visual variability.
Finally, although CLIP and DINOv3 provide strong semantic and patch-level encodings, some results may reflect limitations of the encoders rather than those of the diffusion models. Despite these limitations, our findings suggest that diffusion models should be studied not only in terms of what they reproduce, but also how they transform iconic content, moving beyond simplified views of memorization toward a more nuanced understanding of generative models as systems that encode, reinterpret, and reshape elements of collective visual culture. 

%
%
\bibliographystyle{splncs04}
\bibliography{main}

\clearpage
\appendix
\setcounter{page}{1}

\section{Dataset Composition}
\label{sec:appendix_dataset}

\begin{figure}[h!]
    \centering
    \includegraphics[width=0.9\linewidth]{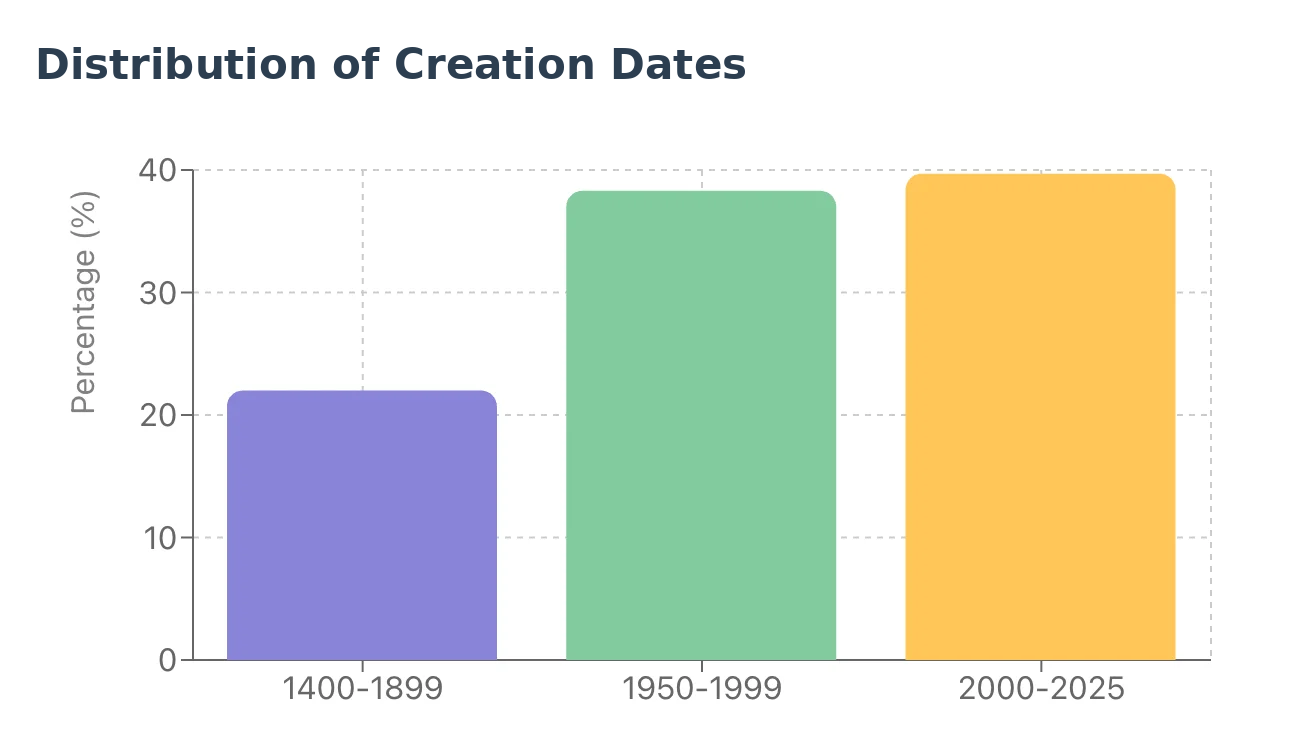}
    \vspace{-2mm}
    \caption{\textbf{Creation Date.} Most concepts originate from the mid- and late-twentieth century onward, providing a modern cultural focus while maintaining historical coverage.}
    \label{fig:supp_temporal}
\end{figure}

The creation date (\cref{fig:supp_temporal}) shows that the majority of concepts originate from the mid- and late-twentieth century onward (1950--1999: 34.1\%; 2000--2025: 39.7\%), while earlier periods (1400--1899: $\approx$22\%) remain proportionally represented, providing a historical baseline for evaluating long-term cultural representation. 

\begin{figure}[h!]
    \centering
    \includegraphics[width=0.9\linewidth]{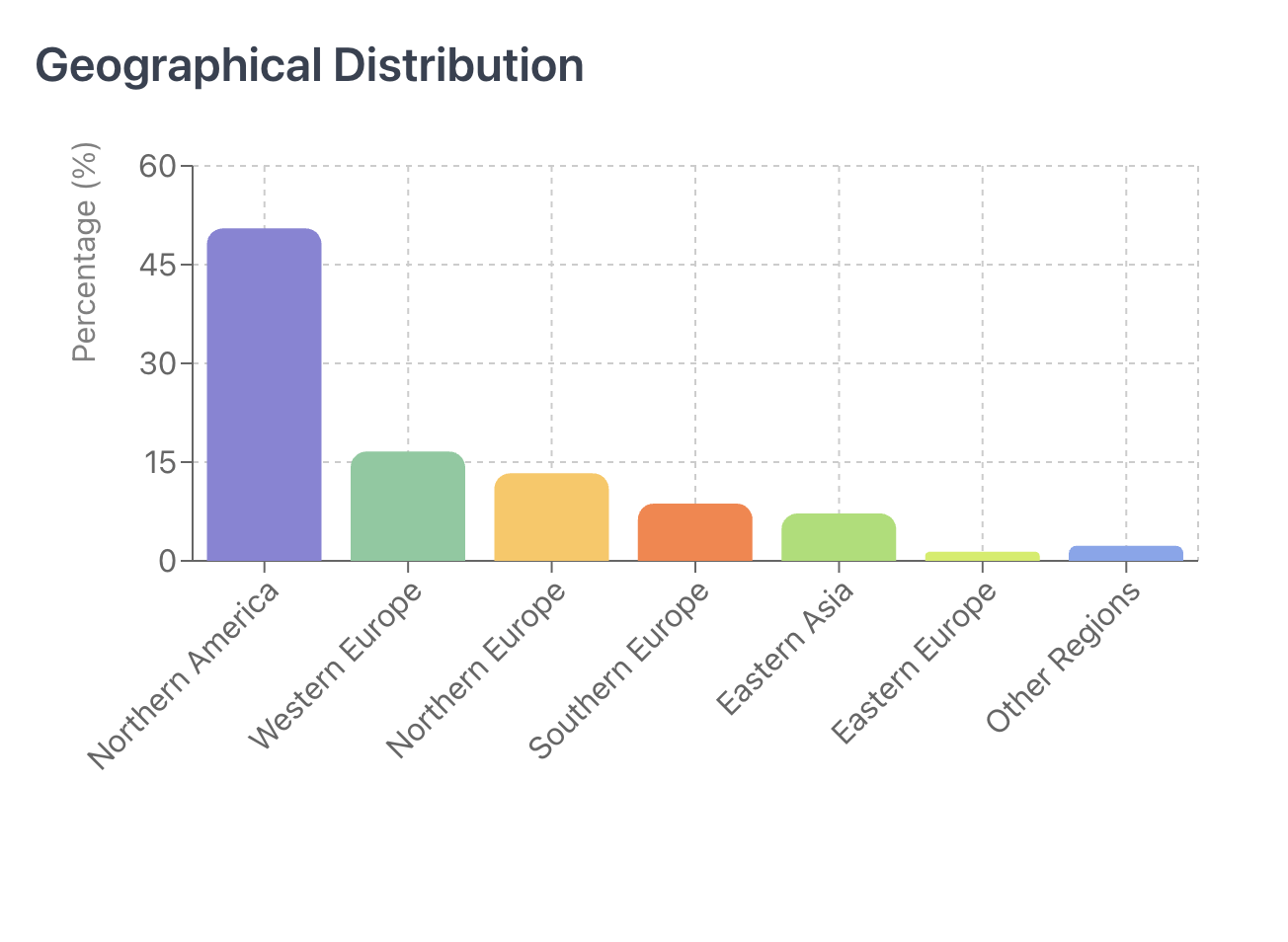}
    \vspace{-2mm}
    \caption{\textbf{Geographical Distribution.} The dataset primarily reflects Northern American and Western European contexts, with contributions from other regions providing broader cultural diversity.}
    \label{fig:supp_geo}
\end{figure}

Geographically (\cref{fig:supp_geo}), the dataset is composed primarily of concepts associated with Northern America (50.5\%) and Western Europe (16.6\%), followed by Northern Europe (13.3\%), Southern Europe (8.7\%), and smaller but non-negligible proportions from Eastern Asia (7.2\%), Eastern Europe (1.4\%), and other regions ($\approx$2\%). This composition reflects the Western focus of the source data while still incorporating globally distributed material.

\begin{figure}[h!]
    \centering
    \includegraphics[width=0.9\linewidth]{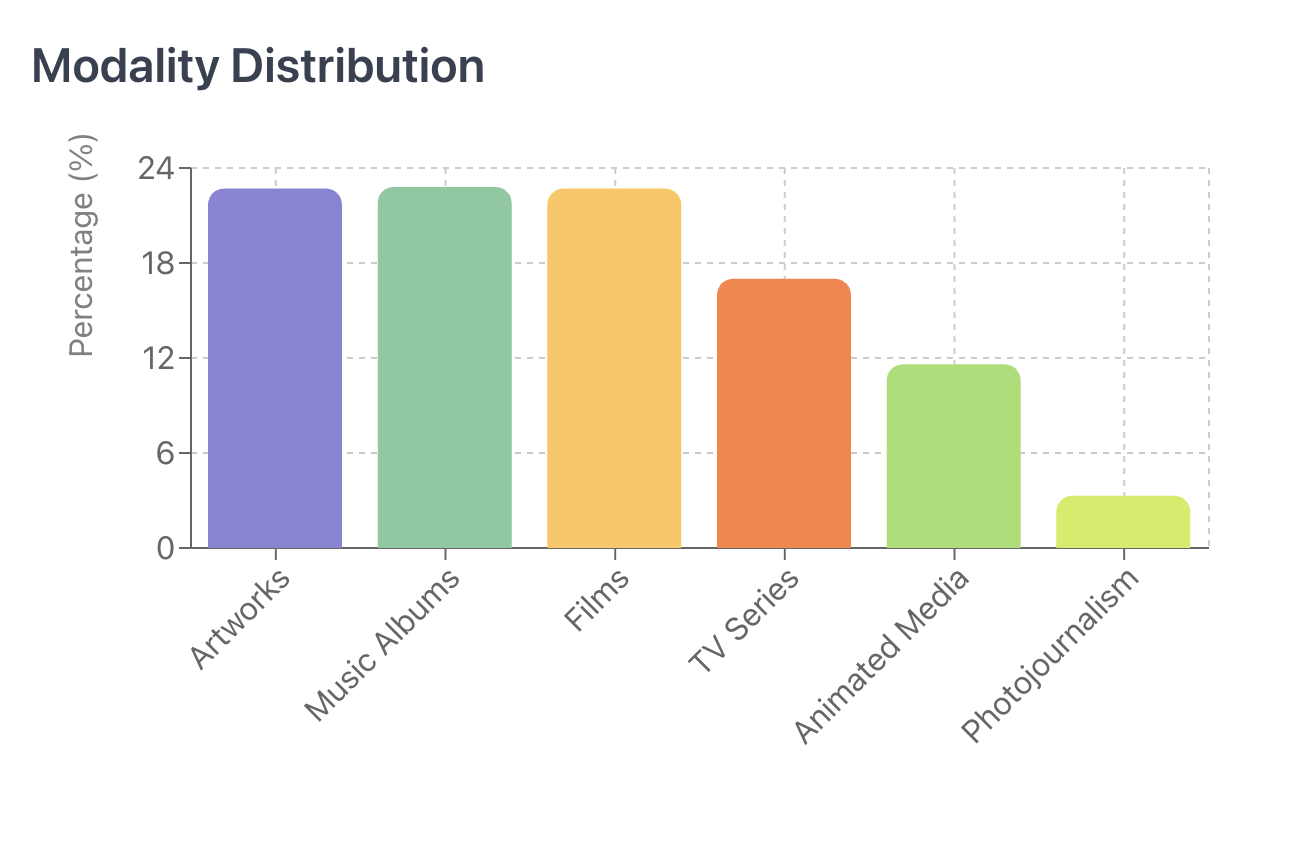}
    \vspace{-2mm}
    \caption{\textbf{Modality Distribution.} Concepts span six major cultural domains, ensuring balanced representation across artistic and media forms.}
    \label{fig:supp_modality}
\end{figure}

In terms of modality (\cref{fig:supp_modality}), the dataset is evenly distributed across major cultural domains, artworks (22.7\%), music albums (22.8\%), films (22.7\%), and television series (17.0\%), complemented by animated media (11.6\%) and photojournalism (3.3\%). Together, these distributions provide a balanced foundation for analyzing how text-to-image models interpret culturally iconic concepts across time, geography, and medium.

\clearpage
\section{Image Generation}
For each cultural reference, each model generates $n=10$ images using the reference title as the prompt, with different random seeds. We generate images at a resolution of 1024×1024 across all models to ensure consistency. For Imagen 4, we use Replicate to make API calls. For FLUX.1-schnell, we set the guidance scale to 3.5, max\_sequence\_length to 256, and use 4 inference steps. For SD3, we use a guidance scale of 4.5 and 30 inference steps. For SDXL, we use a guidance scale of 7.5 and 30 inference steps. For SD2, we use a guidance scale of 7.5 and 50 inference steps. In all cases we use the default scheduler as specified in the respective official model cards.

\section{Threshold Calibration Details}
\label{sec:thresholds}

We empirically determined the thresholds used for recognition and reuse to balance precision and recall when identifying culturally aligned or replicated content.

\begin{figure}[ht!]
    \centering
    \includegraphics[width=0.95\linewidth]{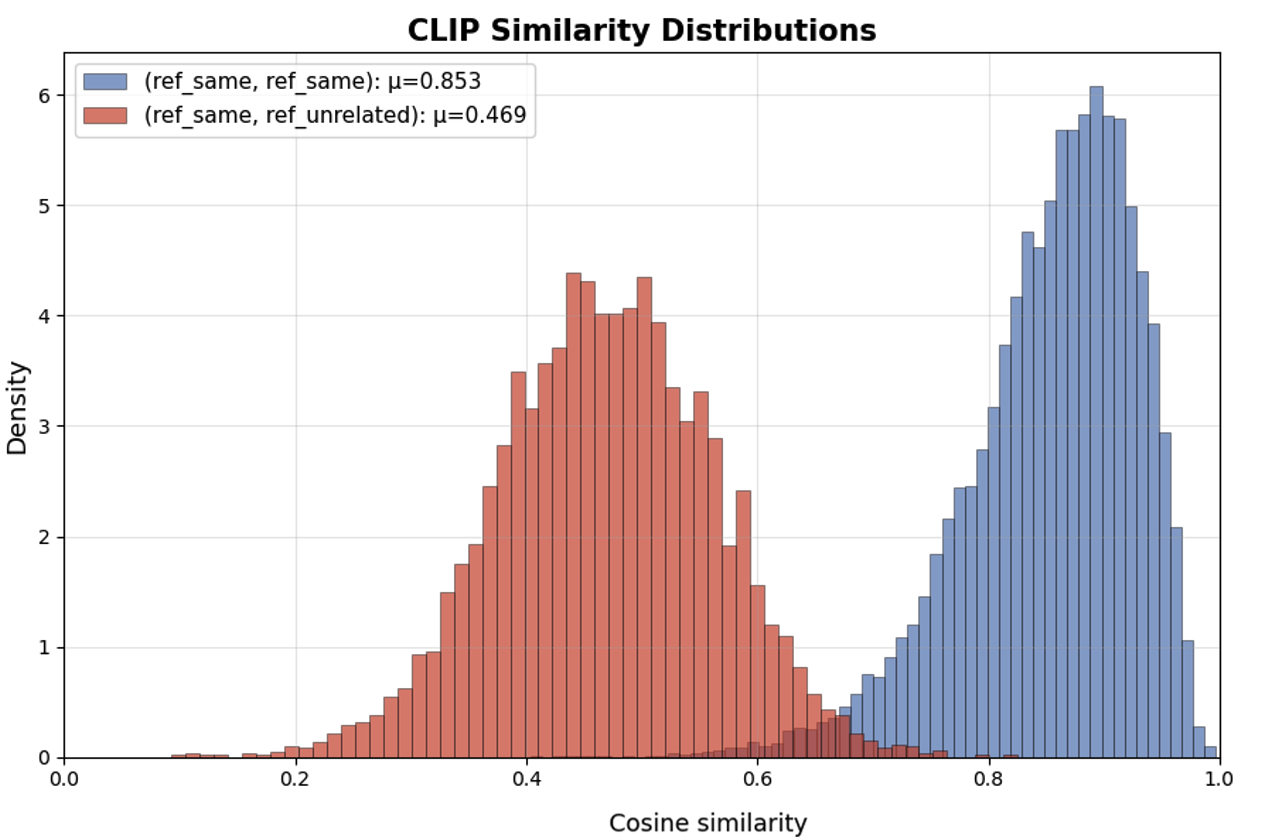}
    \caption{
        \textbf{CLIP similarity distributions.}
        Cosine similarities between reference images of the same (blue) and different (red) cultural references. 
        The separation supports the choice of $\tau=0.7$ for recognition alignment.
    }
    \label{fig:clip_threshold}
\end{figure}

\paragraph{Recognition threshold ($\tau=0.7$).}
To calibrate the recognition threshold, we compared CLIP-based cosine similarities between reference images belonging to the same cultural reference versus different ones. The resulting distributions were well separated ($\mu_{\text{same}} \approx 0.85$, $\mu_{\text{diff}} \approx 0.47$). Setting $\tau=0.7$ retained approximately 96\% of true matches while keeping false positives below 1\%, ensuring that only genuinely related visual pairs were considered aligned (see Fig.~\ref{fig:clip_threshold}).

\begin{figure}[ht!]
    \centering
    \includegraphics[width=0.95\linewidth]{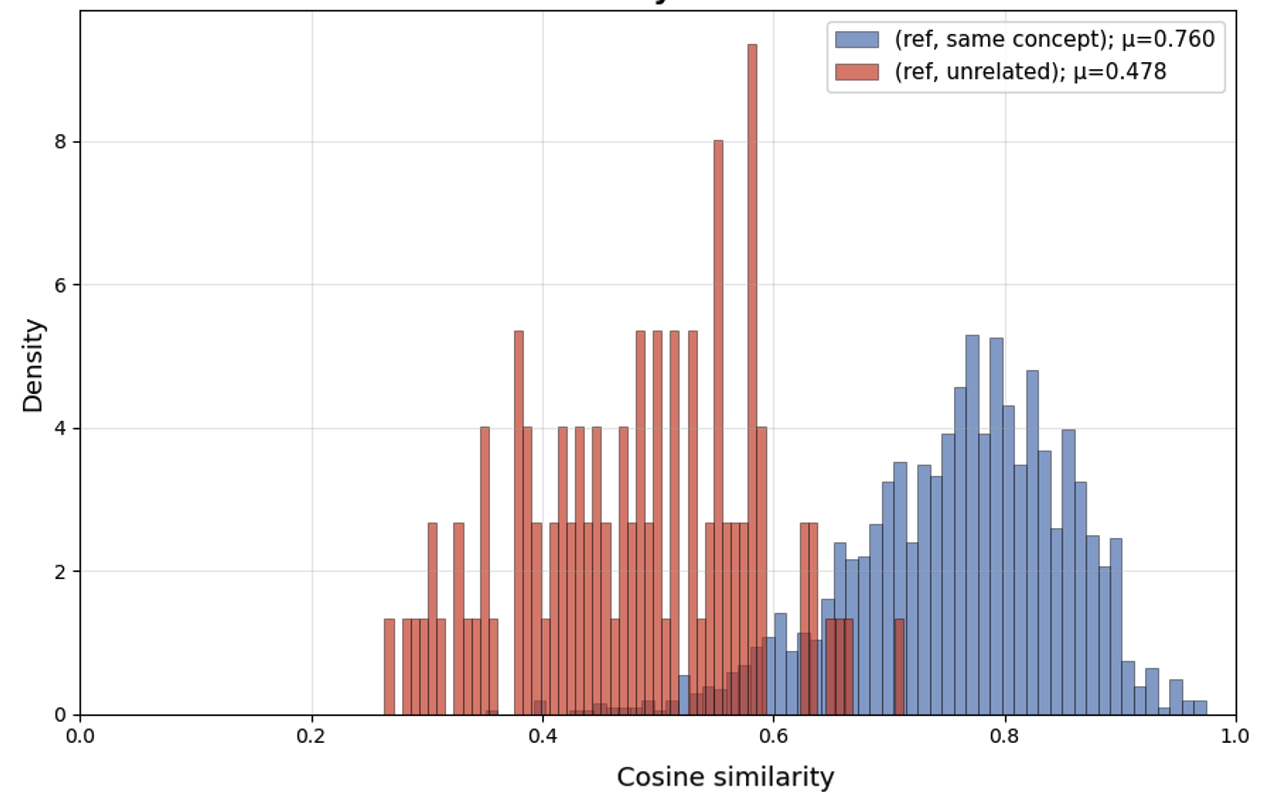}
    \caption{
        \textbf{DINOv3 similarity distributions.}
        Cosine similarities between reference images of the same (blue) and unrelated (red) cultural references. 
        The observed separation motivates the reuse threshold $\tau_{\text{reuse}}=0.6$ used for patch-level analysis.
    }
    \label{fig:dino_threshold}
\end{figure}

\paragraph{Visual Reuse threshold ($\tau_{\text{reuse}}=0.6$).}
For patch-level reuse detection, we analyzed DINOv3 patch similarities between reference images of the same versus unrelated cultural references. Intra-reference similarities averaged $\mu_{\text{same}}\approx0.79$, while unrelated pairs averaged $\mu_{\text{unrelated}}\approx0.48$. Setting $\tau_{\text{reuse}}=0.6$ achieved $F1\approx0.98$ with a false-positive rate below 0.1, effectively distinguishing local replication from unrelated variation. This value aligns with the replication threshold reported by \cite{somepalli2023diffusion} for DINO ($\tau_{\text{reuse}}\approx0.5$), adjusted upward to reflect DINOv3’s finer feature discrimination (see Fig.~\ref{fig:dino_threshold}).

\newpage
\section{Synthetic Validation of Patch-Level VR}
\label{sec:VR_validation}
To evaluate how different replication measures respond to controlled levels of visual overlap, we compared the behavior of PDFE, SSCD, and patch-level VR across systematically constructed overlap conditions. Using 100 still-image cultural references from our example set, we generated 10 comparison pairs per reference spanning four validated levels of replication, then averaged values per reference to obtain $N = 100$ independent observations per condition.

\begin{table}[h!]
\centering
\small
\caption{Synthetic validation of patch-level Visual Reuse (VR) against existing replication metrics.}
\begin{tabular}{lcccccc}
\toprule
& \multicolumn{2}{c}{\textbf{VR}} & \multicolumn{2}{c}{\textbf{SSCD}} & \multicolumn{2}{c}{\textbf{PDFE}} \\
\cmidrule(lr){2-3}\cmidrule(lr){4-5}\cmidrule(lr){6-7}
\textbf{Test Scenario} & Mean$\pm$SD & Min–Max & Mean$\pm$SD & Min–Max & Mean$\pm$SD & Min–Max \\
\midrule
Exact copy        & $0.97\pm0.01$ & 0.94–1.00 & $0.95\pm0.02$ & 0.94–0.98 & $4.7\pm0.06$ & 4.0–5.0 \\
50\% spatial      & $0.51\pm0.03$ & 0.45–0.57 & $0.63\pm0.15$ & 0.41–0.71 & $2.9\pm0.89$ & 0.0–4.0 \\
25\% localized    & $0.27\pm0.05$ & 0.20–0.34 & $0.22\pm0.12$ & 0.16–0.48 & $1.4\pm0.67$ & 0.0–3.0 \\
Unrelated         & $0.02\pm0.02$ & 0.00–0.08 & $0.04\pm0.03$ & 0.00–0.11 & $1.0\pm0.13$ & 0.0–2.0 \\
\bottomrule
\end{tabular}
\label{tab:vr_validation}
\end{table}

Specifically, \textit{(i)} in the \textbf{exact copy} condition (100\% overlap), each reference image was compared to itself ten times to assess metric stability;
\textit{(ii)} in the \textbf{50\% spatial overlap} condition, we generated ten synthetic composites per reference by copying half of the patch grid, either the top or bottom half, or the left or right half, into ten distinct target references from the same set, thereby preserving contiguous spatial correspondence across half the image area;
\textit{(iii)} in the \textbf{25\% localized overlap} condition, 25\% of patches were randomly selected from each reference and inserted as 2$\times$2 blocks into random locations within ten different target references, preserving local spatial coherence (e.g., facial regions or object fragments) while distributing copied content throughout the composition; and \textit{(iv)} in the \textbf{unrelated pair} condition (0\% overlap), each reference was compared to ten different references with no intentional spatial correspondence. This controlled setup, with multiple realizations per reference averaged prior to analysis, enables a fine-grained assessment of how VR, SSCD, and PDFE respond to systematically varied visual overlap, while accounting for variance introduced by target pairings and patch configurations.

As shown in Table~\ref{tab:vr_validation}, VR scales proportionally with the true degree of visual reuse, with mean values of 0.97, 0.51, 0.27, and 0.02 across the exact copy, 50\%, 25\%, and unrelated conditions, respectively. The narrow standard deviations and tight min–max ranges across all levels indicate consistent responses across diverse source–target pairings, even when reused content is localized or dispersed throughout the composition. By contrast, SSCD saturates under exact copying but shows broader variation in intermediate conditions, particularly for 50\% reuse, suggesting that it is less sensitive to structured partial replication. PDFE captures the expected ordinal trends in the mean but exhibits substantial dispersion at both 50\% and 25\% reuse levels (SDs of 0.89 and 0.67), with predictions spanning multiple replication categories. While both SSCD and PDFE remain informative for identifying replication in a broader sense, these results show that VR better quantifies the extent of visual reuse, especially when localized replication coexists with compositional variability.

\section{Variation Within PDFE Levels}
\label{sec:pdfe_variation}
To examine how the disentangled evaluation captures distinct aspects of multimodal iconicity, we analyzed the relationship between CRA, VR, and CRT within PDFE replication levels across all evaluated models. As shown in Table~\ref{tab:variance_pdfe}, at intermediate PDFE levels (2–4) both CRA and CRT display wide dispersion, with standard deviations between 0.25 and 0.40 for CRA and between 0.23 and 0.30 for CRT, spanning the full [0,1] range across still and moving-image references alike. Even at high replication levels (PDFE=4–5), VR values remain highly variable—for example, between 0.04 and 0.93 at PDFE=5, indicating that perceptual similarity does not systematically correspond to visual reuse. These findings reveal that discrete replication levels conflate mechanistically distinct generation strategies, ranging from learned cultural transformation to near-exact copying.

\begin{table}[ht!]
\centering
\footnotesize
\caption{Summary statistics of CRA, VR, and CRT across PDFE replication levels for still- and moving-image references. Values report mean $\pm$ SD, minimum--maximum range, and sample size ($n$).}
\label{tab:variance_pdfe}

\resizebox{\linewidth}{!}{%
\begin{tabular}{@{}l *{3}{ccc} @{}}
\toprule
& \multicolumn{3}{c}{\textbf{CRA}}
& \multicolumn{3}{c}{\textbf{VR}}
& \multicolumn{3}{c}{\textbf{CRT}} \\
\cmidrule(lr){2-4}\cmidrule(lr){5-7}\cmidrule(lr){8-10}
\textbf{PDFE}
& $\mu \pm \sigma$ & Min--Max & $n$
& $\mu \pm \sigma$ & Min--Max & $n$
& $\mu \pm \sigma$ & Min--Max & $n$ \\
\midrule

\multicolumn{10}{@{}l}{\textit{Still-image References}} \\
\addlinespace[2pt]

0 & 0.08$\pm$0.20 & 0.0--1.0 & 147 & 0.04$\pm$0.14 & 0.0--0.84 & 147 & 0.33$\pm$0.26 & 0.03--1.0 & 147 \\
1 & 0.12$\pm$0.26 & 0.0--1.0 & 511 & 0.10$\pm$0.16 & 0.0--0.88 & 511 & 0.37$\pm$0.27 & 0.0--1.0 & 511 \\
2 & 0.30$\pm$0.38 & 0.0--1.0 & 674 & 0.17$\pm$0.20 & 0.0--0.95 & 674 & 0.45$\pm$0.28 & 0.04--1.0 & 674 \\
3 & 0.66$\pm$0.40 & 0.0--1.0 & 396 & 0.27$\pm$0.23 & 0.0--1.0 & 396 & 0.55$\pm$0.26 & 0.0--1.0 & 396 \\
4 & 0.86$\pm$0.29 & 0.0--1.0 & 93  & 0.36$\pm$0.26 & 0.0--0.99 & 93  & 0.56$\pm$0.25 & 0.01--1.0 & 93 \\
5 & 0.95$\pm$0.13 & 0.4--1.0 & 24  & 0.49$\pm$0.28 & 0.04--0.93 & 24  & 0.47$\pm$0.25 & 0.08--0.9 & 24 \\

\midrule

\multicolumn{10}{@{}l}{\textit{Moving-image References}} \\
\addlinespace[2pt]

0 & 0.13$\pm$0.22 & 0.0--0.7 & 15  & 0.03$\pm$0.05 & 0.0--0.15 & 15  & 0.11$\pm$0.20 & 0.0--0.63 & 15 \\
1 & 0.20$\pm$0.30 & 0.0--1.0 & 111 & 0.08$\pm$0.15 & 0.0--0.81 & 111 & 0.16$\pm$0.24 & 0.0--0.94 & 111 \\
2 & 0.57$\pm$0.37 & 0.0--1.0 & 493 & 0.15$\pm$0.16 & 0.0--0.83 & 493 & 0.46$\pm$0.30 & 0.0--0.99 & 493 \\
3 & 0.74$\pm$0.31 & 0.0--1.0 & 864 & 0.23$\pm$0.19 & 0.0--0.99 & 864 & 0.55$\pm$0.26 & 0.0--0.99 & 864 \\
4 & 0.85$\pm$0.25 & 0.0--1.0 & 397 & 0.28$\pm$0.18 & 0.0--0.96 & 397 & 0.60$\pm$0.23 & 0.0--0.98 & 397 \\
5 & 0.89$\pm$0.21 & 0.1--1.0 & 80  & 0.30$\pm$0.18 & 0.0--0.93 & 80  & 0.61$\pm$0.21 & 0.08--1.0 & 80 \\

\bottomrule
\end{tabular}%
}
\end{table}

\begin{figure*}[ht!]
\centering
\includegraphics[width=\textwidth]{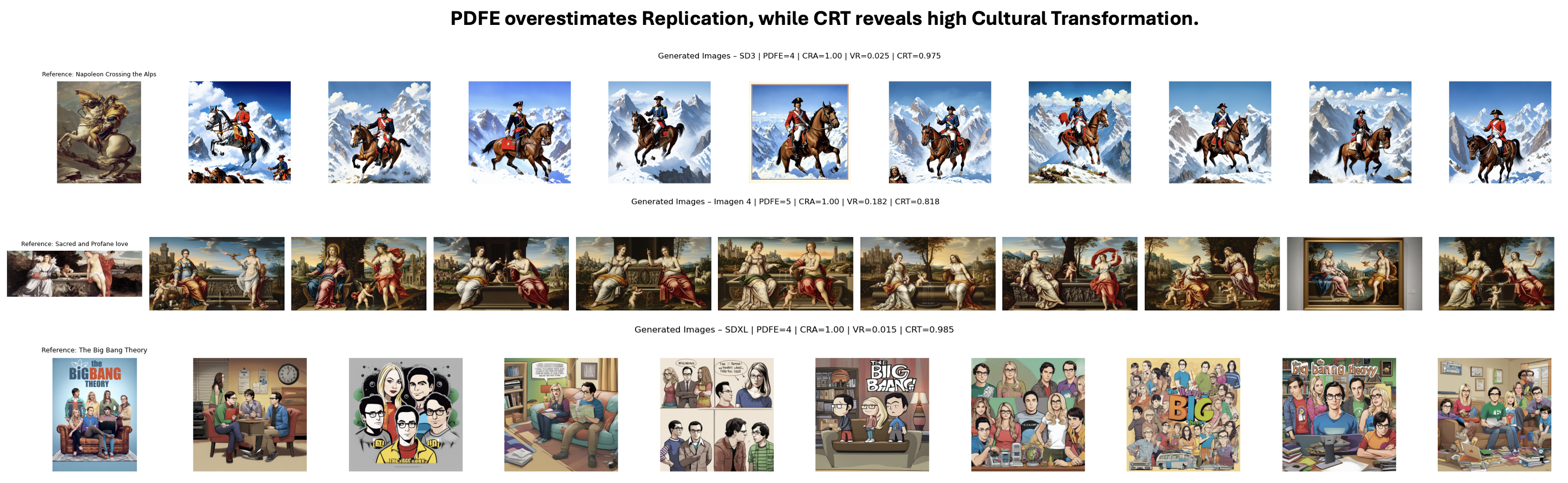}
\caption{\textbf{PDFE overestimates replication while CRT reveals cultural transformation.}
Compositional coherence stems from learned iconic structure rather than direct visual reuse (\textit{Napoleon Crossing the Alps}, \textit{Sacred and Profane Love}, \textit{The Big Bang Theory}).}
\label{fig:pdfe_overestimates}
\end{figure*}

\begin{figure*}[ht!]
\centering
\includegraphics[width=\textwidth]{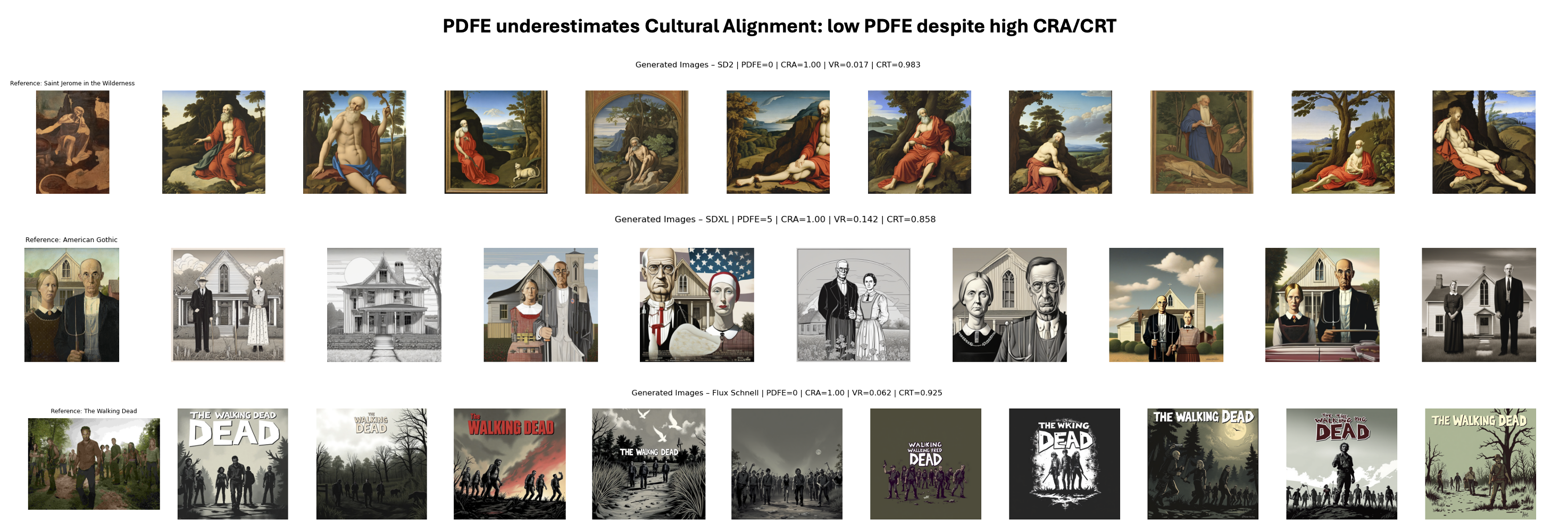}
\caption{\textbf{PDFE underestimates cultural alignment:} low PDFE despite high CRA/CRT.
Models reproduce canonical iconography through transformation rather than replication (\textit{Saint Jerome in the Wilderness}, \textit{American Gothic}, \textit{The Walking Dead}).}
\label{fig:pdfe_underestimates}
\end{figure*}

Figures~\ref{fig:pdfe_underestimates}–\ref{fig:pdfe_overestimates} illustrate these differences through representative examples. In several cases, PDFE \textit{underestimates} cultural alignment, assigning low replication scores to generations that accurately reproduce canonical iconography without reusing visual material, as shown in Figure~\ref{fig:pdfe_underestimates}. In \textit{Saint Jerome in the Wilderness} (SD2) and \textit{American Gothic} (SDXL), models achieve high CRA and low VR, capturing the compositional and symbolic essence of the reference while remaining visually independent. Similarly, in \textit{The Walking Dead} (Flux Schnell), the model achieves full recognition through stylistic transformation of the ensemble silhouette rather than reuse of specific imagery.

In other cases, PDFE \textit{overestimates} replication when compositional coherence arises from learned iconic structure rather than direct visual reuse, as illustrated in Figure~\ref{fig:pdfe_overestimates}. Generations of \textit{Napoleon Crossing the Alps} (SD3) and \textit{Sacred and Profane Love} (Imagen~4) receive high replication scores despite minimal VR: the models reproduce canonical arrangements while varying perspective, detail, and style. A similar pattern appears for \textit{The Big Bang Theory} (SDXL), where visually diverse renderings share only the recognizable ensemble composition, yielding high CRA and CRT but negligible VR.

\begin{figure*}[ht!]
\centering
\includegraphics[width=\textwidth]{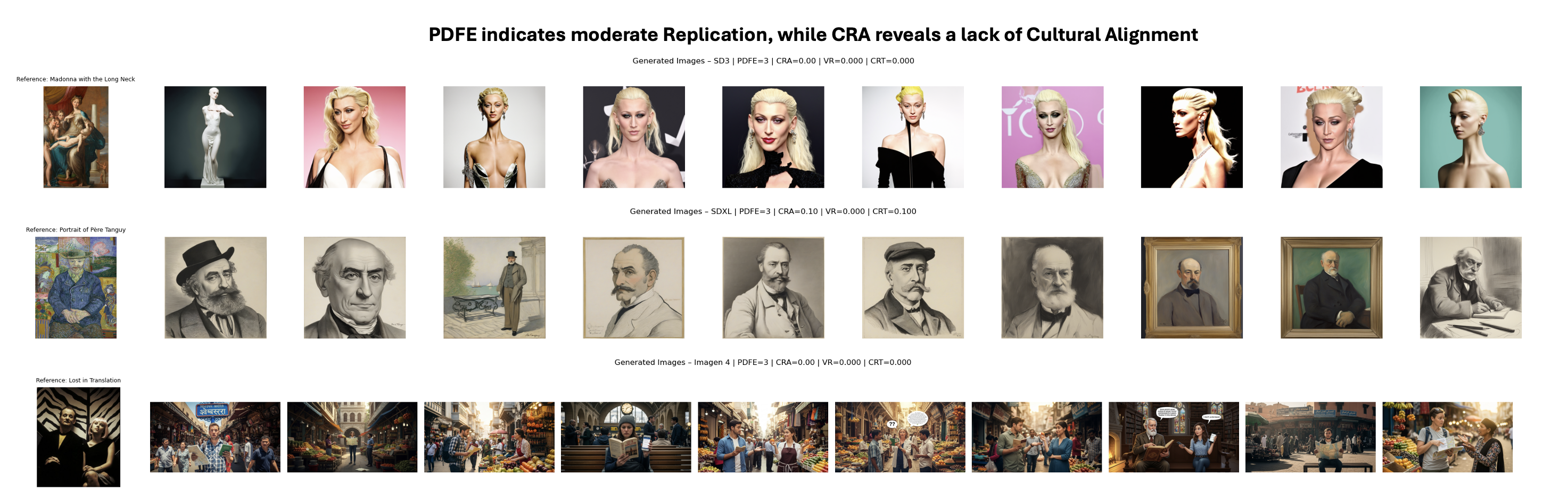}
\caption{\textbf{PDFE indicates moderate replication while CRA reveals lack of cultural alignment.}
Models generate superficially similar scenes without capturing the intended cultural reference (\textit{Madonna with the Long Neck}, \textit{Portrait of Père Tanguy}, \textit{Lost in Translation}).}
\label{fig:pdfe_moderate}
\end{figure*}

Finally, moderate PDFE scores can conceal \textit{cultural misalignment}, where generations appear perceptually similar yet fail to capture the intended iconic relationship between text and image, as shown in Figure~\ref{fig:pdfe_moderate}. In these cases, models produce literal depictions of the prompt rather than culturally grounded interpretations. For instance, \textit{Madonna with the Long Neck} (SD3) and \textit{Portrait of Père Tanguy} (SDXL) yield generic portraits that omit the defining iconography and stylistic cues of their respective references, resulting in low CRA and CRT despite mid-level replication predictions. Likewise, \textit{Lost in Translation} (Imagen~4) produces generic urban scenes that reflect a literal interpretation of the prompt rather than evoking the film’s iconic visual motifs and atmosphere.

\clearpage
\section{Human Evaluation Study}
\label{sec:human_evaluation}

\begin{figure}[ht!]
    \centering
    \includegraphics[width=\linewidth]{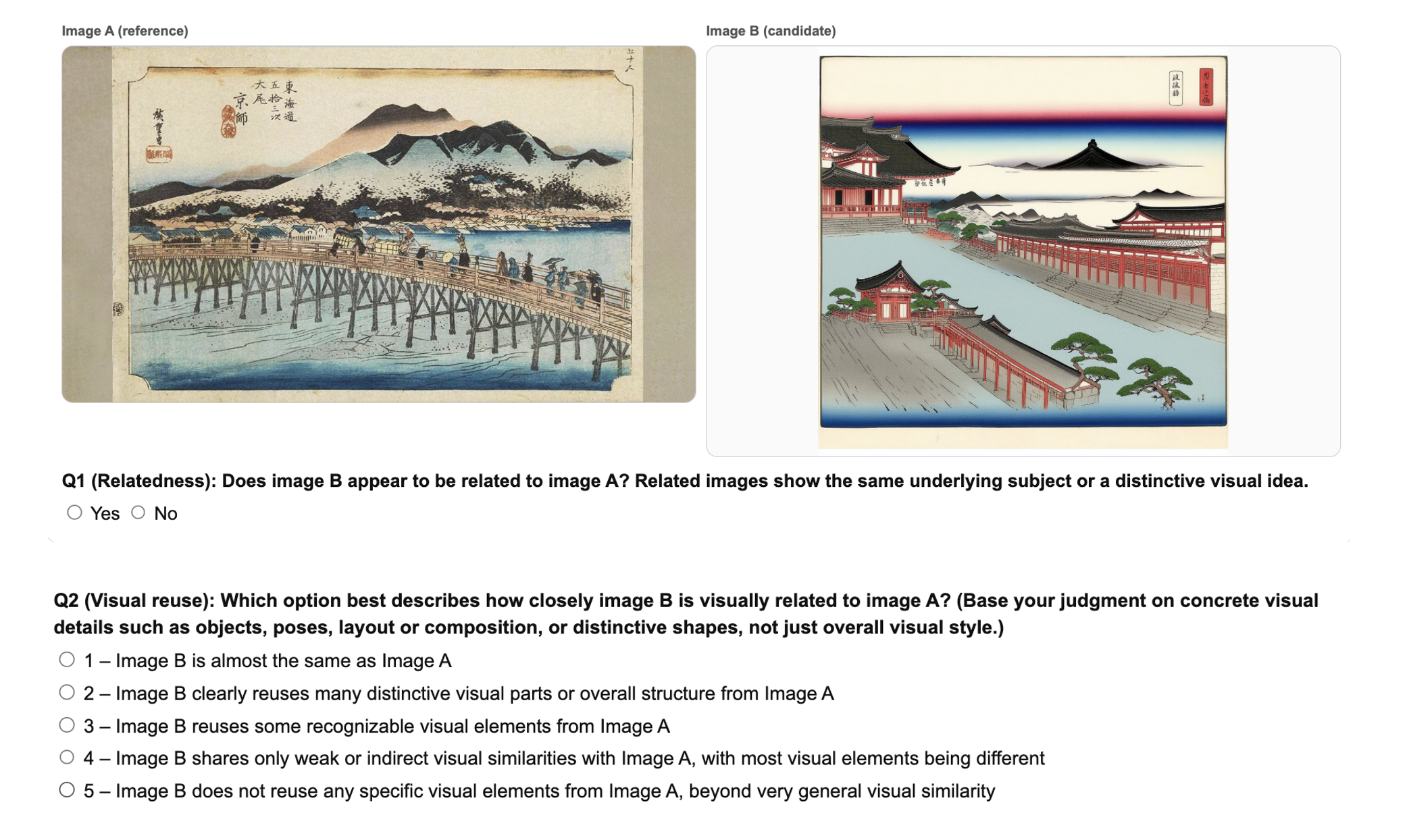}
    \caption{Example annotation item from the human evaluation interface. Annotators were shown a reference image and a candidate image side by side, then asked to judge relatedness (Q1) and, when applicable, the degree of visual reuse (Q2).}
    \label{fig:user-study-example-question}
\end{figure}

\begin{figure}[ht!]
    \centering
    \includegraphics[width=\linewidth]{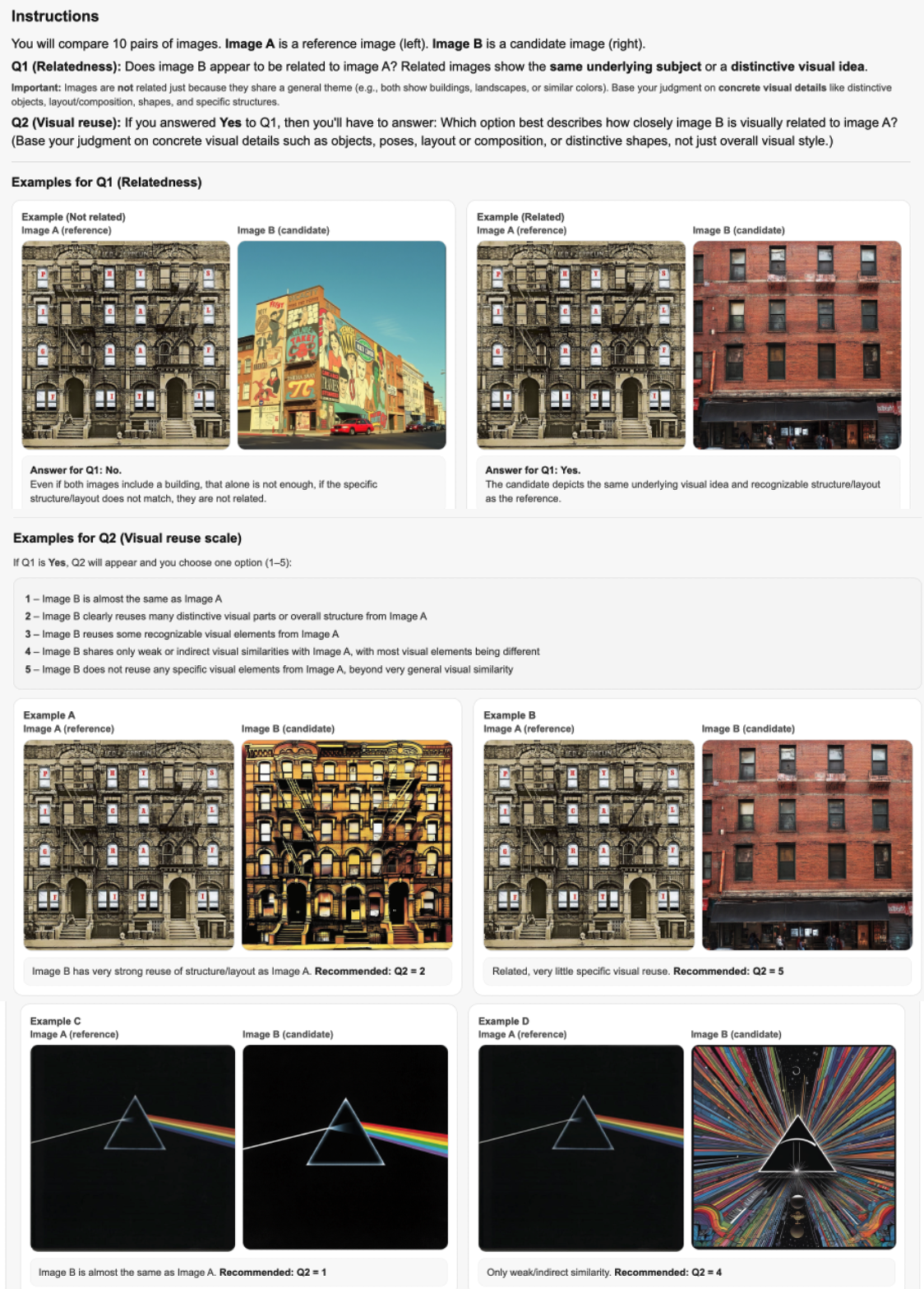}
    \caption{Instruction and calibration screen shown before annotation. The interface explains the relatedness judgment (Q1), the five-point visual reuse scale (Q2), and provides example image pairs illustrating the intended distinctions.}
    \label{fig:user-study-instructions}
\end{figure}

Figures \ref{fig:user-study-instructions} and \ref{fig:user-study-example-question} show the annotation interface used in the human evaluation study. Annotators were presented with a reference image (\textit{Image A}) and a candidate generated image (\textit{Image B}) and asked to complete two questions. First, they answered a binary \textbf{relatedness} question (Q1): whether the candidate image appeared to evoke the same underlying cultural reference or distinctive visual idea as the reference image. Annotators were instructed to base this judgment on concrete visual evidence, such as distinctive objects, layout, composition, poses, or structural elements, rather than broad thematic similarity alone. If annotators answered \textit{Yes} to Q1, they then answered a \textbf{visual reuse} question (Q2), rating how closely the candidate reused specific visual content from the reference on a five-point scale. The scale ranged from near-identity (\textit{Image B is almost the same as Image A}) to no specific visual reuse beyond general similarity.

To calibrate judgments, the interface included several illustrative examples for both Q1 and Q2 before the task began. \Cref{fig:user-study-instructions} shows the instruction and calibration screen presented to participants, and \Cref{fig:user-study-example-question} shows an example evaluation item from the annotation interface.

\clearpage
\section{Model-Level Statistics}
\label{sec:appendix_model_stats}
To complement the main model-level comparison, we report the full aggregate statistics underlying \cref{tab:static_dynamic_vr_vrk_crtbin}, together with pairwise statistical comparisons of model differences in VR, VR@k and CRT. The subsequent tables report pairwise comparisons between models based on per-reference values, including the mean difference, bootstrap 95\% confidence interval, and Holm-corrected \(p\)-value from a Wilcoxon signed-rank test. 

\begin{table}[ht!]
\centering
\caption{Pairwise statistical significance results for still-image references. Differences (\textit{Diff}) represent model 1 minus model 2. $P$-values are bolded where significant ($p < 0.05$) after Holm-Bonferroni correction.}
\label{tab:appendix_pairwise_still}
\scriptsize
\setlength{\tabcolsep}{4.5pt}
\renewcommand{\arraystretch}{1.1}
\begin{tabular}{lcccccc}
\toprule
& \multicolumn{2}{c}{\textbf{CRT}} & \multicolumn{2}{c}{\textbf{Shared VR (VR@k)}} & \multicolumn{2}{c}{\textbf{Global VR (G-VR)}} \\
\cmidrule(lr){2-3} \cmidrule(lr){4-5} \cmidrule(lr){6-7}
\textbf{Comparison} & \textbf{Diff} & \textbf{$P$-value} & \textbf{Diff} & \textbf{$P$-value} & \textbf{Diff} & \textbf{$P$-value} \\
\midrule
Flux Schnell vs Imagen 4 & -0.090 & \textbf{0.0009} & -0.137 & \textbf{<0.0001} & -0.173 & \textbf{<0.0001} \\
Flux Schnell vs SD2      & -0.003 & 0.9489          & -0.150 & \textbf{<0.0001} & -0.155 & \textbf{<0.0001} \\
Flux Schnell vs SD3      & -0.089 & \textbf{0.0004} & -0.062 & 0.1210           & -0.057 & 0.3469           \\
Flux Schnell vs SDXL     & -0.068 & \textbf{<0.0001} & -0.131 & \textbf{0.001}  & -0.146 & \textbf{<0.0001} 
\\
Imagen 4 vs SD2          &  0.088 & \textbf{0.0001} & -0.013 & 0.7381           &  0.018 & 0.9495           \\
Imagen 4 vs SD3          &  0.002 & 1.0000          &  0.075 & \textbf{0.0481}  &  0.116 & \textbf{<0.0001} 
\\
Imagen 4 vs SDXL         & 0.022 & 0.2532          &  0.005 & 0.9001           &  0.027 & \textbf{0.0285}  \\
SD2 vs SD3               & -0.086 & \textbf{<0.0001} &  0.088 & \textbf{0.0025}  &  0.098 & \textbf{<0.0001} \\
SD2 vs SDXL              & -0.065 & \textbf{<0.001} &  0.019 & 0.3554  &  0.009 & 0.628  \\

SD3 vs SDXL              & 0.021 & \textbf{0.0284}          & -0.069 & \textbf{<0.0001}           & -0.089 & \textbf{0.0004}           \\
\bottomrule
\end{tabular}
\end{table}

\begin{table}[ht!]
\centering
\caption{Pairwise statistical significance results for moving-image references. Differences (\textit{Diff}) represent model 1 minus model 2. $P$-values are bolded where significant ($p < 0.05$) after Holm-Bonferroni correction.}
\label{tab:appendix_pairwise_moving}
\scriptsize
\setlength{\tabcolsep}{4.5pt}
\renewcommand{\arraystretch}{1.1}
\begin{tabular}{lcccccc}
\toprule
& \multicolumn{2}{c}{\textbf{CRT (Binary)}} & \multicolumn{2}{c}{\textbf{Shared VR (VR@k)}} & \multicolumn{2}{c}{\textbf{Global VR (G-VR)}} \\
\cmidrule(lr){2-3} \cmidrule(lr){4-5} \cmidrule(lr){6-7}
\textbf{Comparison} & \textbf{Diff} & \textbf{$P$-value} & \textbf{Diff} & \textbf{$P$-value} & \textbf{Diff} & \textbf{$P$-value} \\
\midrule
Flux Schnell vs Imagen 4 & -0.130 & \textbf{<0.0001} &  0.035 & \textbf{0.0240} &  0.032 & \textbf{0.0342} \\
Flux Schnell vs SD2      & -0.104 & 0.1250          & -0.060 & \textbf{<0.0001} & -0.044 & \textbf{0.0153} \\
Flux Schnell vs SD3      & -0.116 & \textbf{0.0152} & -0.053 & \textbf{<0.0001} & -0.036 & \textbf{0.0153} \\
Flux Schnell vs SDXL     & -0.041 & \textbf{0.0101} &  0.063 & \textbf{<0.0001} &  0.054 & \textbf{<0.0001} \\
Imagen 4 vs SD2          &  0.026 & \textbf{0.0037} & -0.095 & \textbf{<0.0001} & -0.076 & \textbf{<0.0001} \\
Imagen 4 vs SD3          &  0.014 & \textbf{0.0196} & -0.088 & \textbf{<0.0001} & -0.069 & \textbf{<0.0001} \\
Imagen 4 vs SDXL         &  0.090 & 0.1822          &  0.028 & \textbf{0.0037} &  0.022 & \textbf{<0.0001} \\
SD2 vs SD3               & -0.012 & 1.0000          &  0.007 & 0.5286          &  0.008 & 0.8951          \\
SD2 vs SDXL              &  0.064 & 1.0000          &  0.124 & \textbf{<0.0001} &  0.098 & \textbf{<0.0001} \\
SD3 vs SDXL              &  0.076 & 1.0000          &  0.117 & \textbf{<0.0001} &  0.090 & \textbf{<0.0001} \\
\bottomrule
\end{tabular}
\end{table}

\section{Backbone Robustness and Rank-Consistency Analysis}
\label{sec:cra_encoders}
To verify that the model-level conclusions reported in Section 5.2 are not sensitive to the choice of image encoder, we recomputed CRA for all five diffusion models using three alternative encoders: \textit{CLIP ViT-L/14}, \textit{OpenCLIP ViT-H/14}, and \textit{SigLIP}. As shown in Table~\ref{tab:cra_backbone_robustness}, the tier structure identified with the baseline ViT-B/32 is preserved across all encoder choices. On still-image references, \textit{Imagen 4} consistently achieves the highest CRA while \textit{Flux Schnell} consistently trails. On moving-image references, \textit{SD2} and \textit{SD3} consistently form a top tier, separated from \textit{Imagen 4}, \textit{SDXL}, and \textit{Flux Schnell}. While minor rank-swapping occurs between models with near-identical baseline scores (e.g., SD2 and SD3 on moving images), these pairs were already found to be statistically indistinguishable. We observe strong Spearman rank-correlations ($\rho$) ranging from 0.80 to 0.92, confirming that our comparative findings are a property of the generative models rather than an artifact of the ViT-B/32 embedding space.

\begin{table}[ht!]
\centering
\caption{\textbf{CRA Backbone Robustness.} Mean CRA scores across different image encoders for still and moving-image references.}
\label{tab:cra_backbone_robustness}
\small 
\setlength{\tabcolsep}{8pt} 
\renewcommand{\arraystretch}{1.1}
\begin{tabular}{lcccc}
\toprule
\textbf{Model} & \textbf{ViT-B/32} & \textbf{ViT-L/14} & \textbf{ViT-H/14} & \textbf{SigLIP} \\
\midrule
\multicolumn{5}{l}{\textit{Still-Image References}} \\
Flux Schnell & 0.401 & 0.428 & 0.388 & 0.342 \\
Imagen 4     & \textbf{0.623} & \textbf{0.651} & \textbf{0.618} & \textbf{0.574} \\
SD2          & 0.489 & 0.505 & 0.512 & 0.428 \\
SD3          & 0.525 & 0.562 & 0.541 & 0.485 \\
SDXL         & 0.572 & 0.634 & 0.589 & 0.521 \\
\midrule
\multicolumn{5}{l}{\textit{Moving-Image References}} \\
Flux Schnell & 0.679 & 0.702 & 0.655 & 0.601 \\
Imagen 4     & 0.816 & 0.844 & 0.802 & 0.758 \\
SD2          & 0.867 & 0.875 & \textbf{0.884} & 0.814 \\
SD3          & \textbf{0.875} & \textbf{0.890} & 0.871 & \textbf{0.829} \\
SDXL         & 0.684 & 0.715 & 0.668 & 0.635 \\
\bottomrule
\end{tabular}
\end{table}

\section{Model-Level Comparison (Qualitative Examples)}
\label{sec:qualitative_examples}

\begin{figure*}[ht!]
    \centering
    \includegraphics[width=0.8\textwidth]{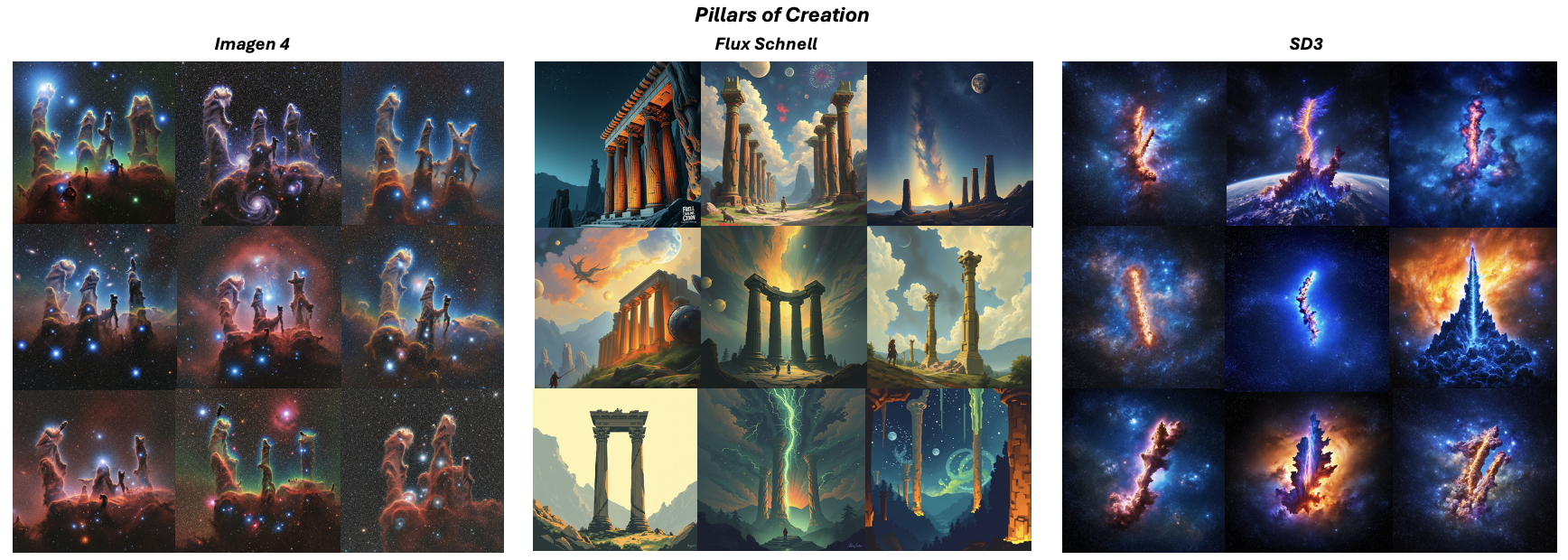}
    \caption{
    \textbf{Prompt:} \emph{Pillars of Creation}. 
    \textbf{Models:} Imagen~4 (left), Flux Schnell (center), SD3 (right). 
    \textbf{CRA:} Imagen~4 = 1.0;\; Flux Schnell = 0.0;\; SD3 = 1.0. 
    \textbf{CRT:} Imagen~4 = 0.00;\; Flux Schnell = 0.00;\; SD3 = 0.73.
    }
    \label{fig:pillars_of_creation}
\end{figure*}

\begin{figure*}[ht!]
    \centering
    \includegraphics[width=0.8\textwidth]{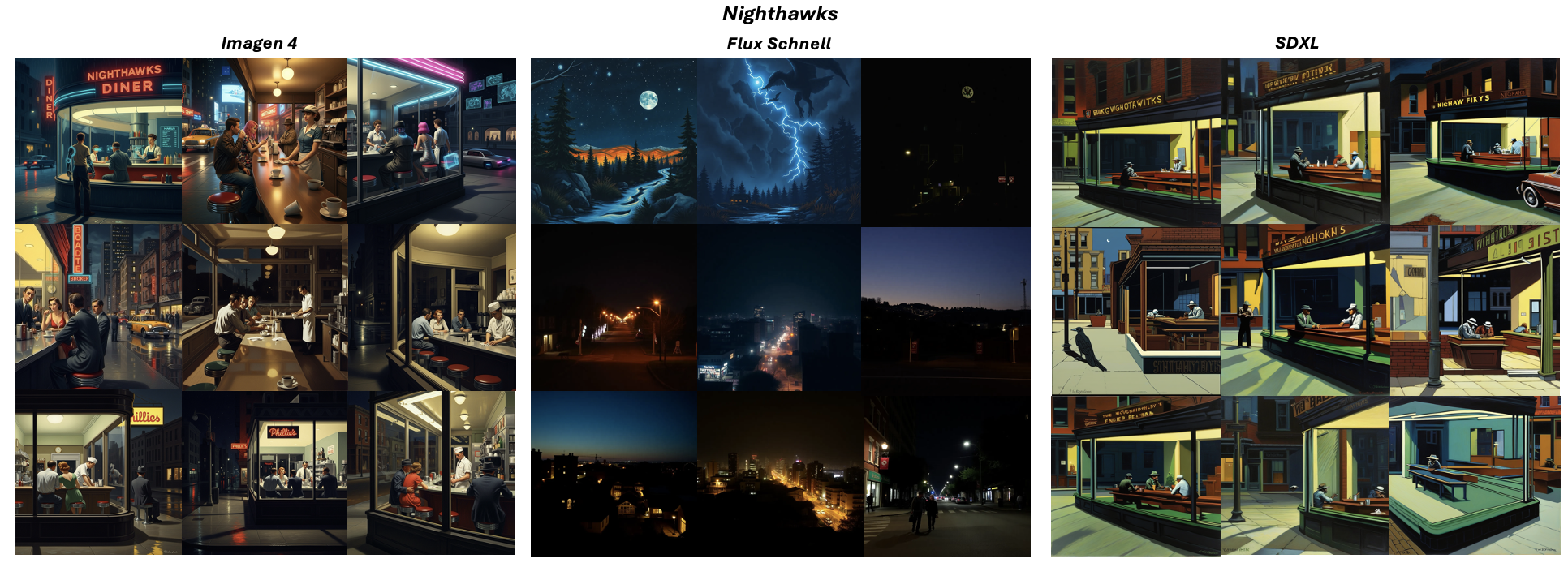}
    \caption{
    \textbf{Prompt:} \emph{Nighthawks}. 
    \textbf{Models:} Imagen~4 (left), Flux Schnell (center), SDXL (right). 
    \textbf{CRA:} Imagen~4 = 1.0;\; Flux Schnell = 0.0;\; SDXL = 1.0. 
    \textbf{CRT:} Imagen~4 = 0.88;\; Flux Schnell = 0.00;\; SDXL = 0.19.
    }
    \label{fig:nighthawks_comparison}
\end{figure*}

\begin{figure*}[ht!]
    \centering
    \includegraphics[width=0.8\textwidth]{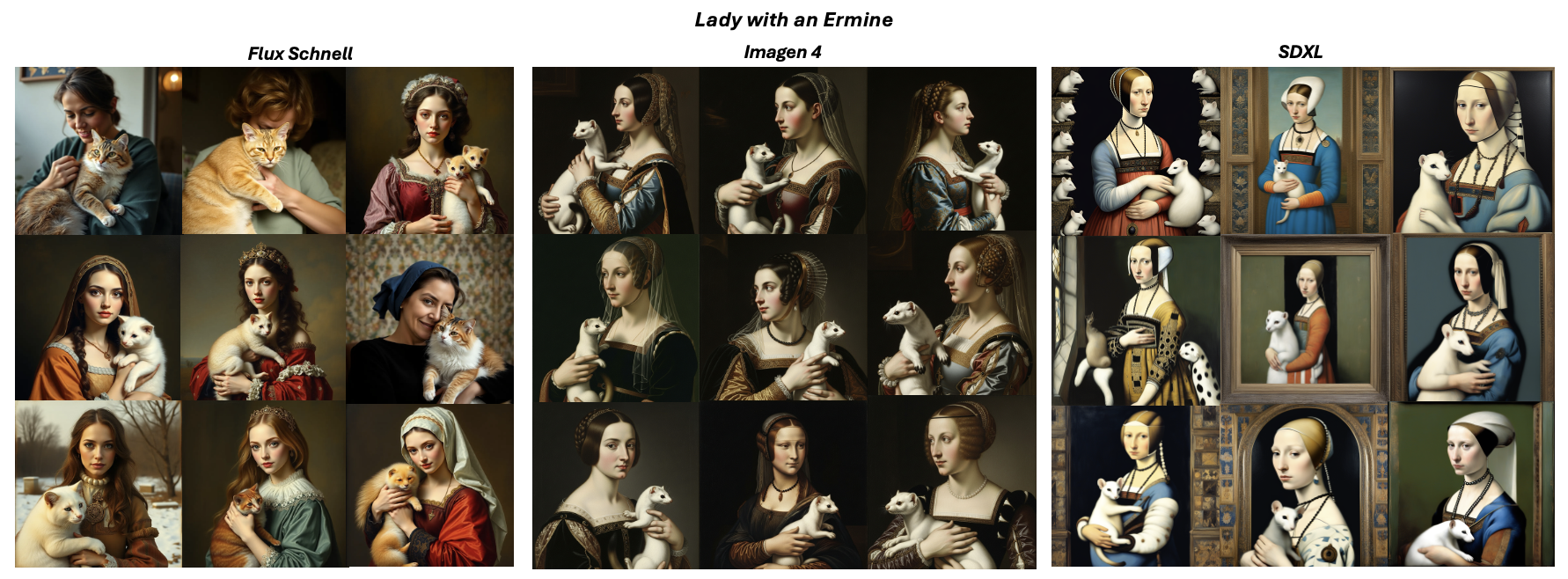}
    \caption{
    \textbf{Prompt:} \emph{Lady with an Ermine}. 
    \textbf{Models:} Flux Schnell (left), Imagen~4 (center), SDXL (right). 
    \textbf{CRA:} Flux Schnell = 0.0;\; Imagen~4 = 1.0;\; SDXL = 1.0. 
    \textbf{CRT:} Flux Schnell = 0.00;\; Imagen~4 = 0.87;\; SDXL = 0.68.
    }
    \label{fig:lady_with_an_ermine}
\end{figure*}

\begin{figure*}[ht!]
    \centering
    \includegraphics[width=0.8\textwidth]{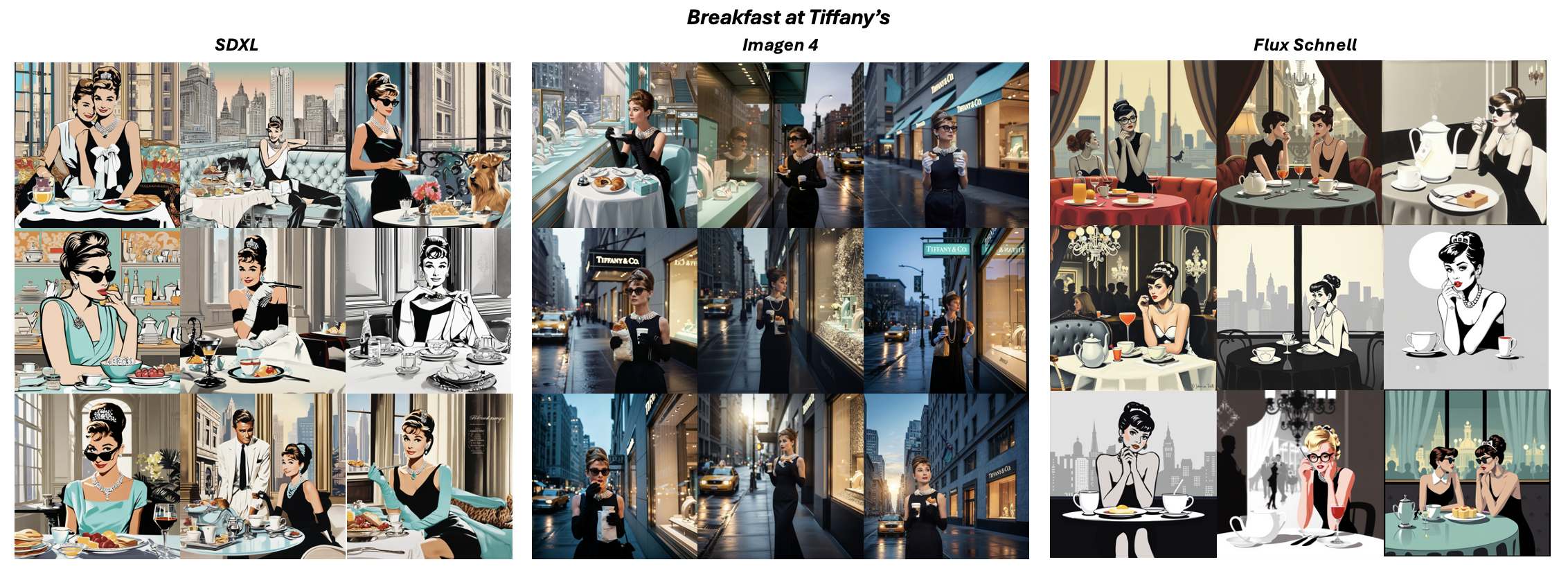}
    \caption{
    \textbf{Prompt:} \emph{Breakfast at Tiffany’s}. 
    \textbf{Models:} SDXL (left), Imagen~4 (center), Flux Schnell (right). 
    \textbf{CRA:} SDXL = 1.0;\; Imagen~4 = 1.0;\; Flux Schnell = 1.0. 
    \textbf{CRT:} SDXL = 0.98;\; Imagen~4 = 0.32;\; Flux Schnell = 0.96.
    }
    \label{fig:breakfast_at_tiffanys}
\end{figure*}

\begin{figure*}[t]
    \centering
    \includegraphics[width=0.8\textwidth]{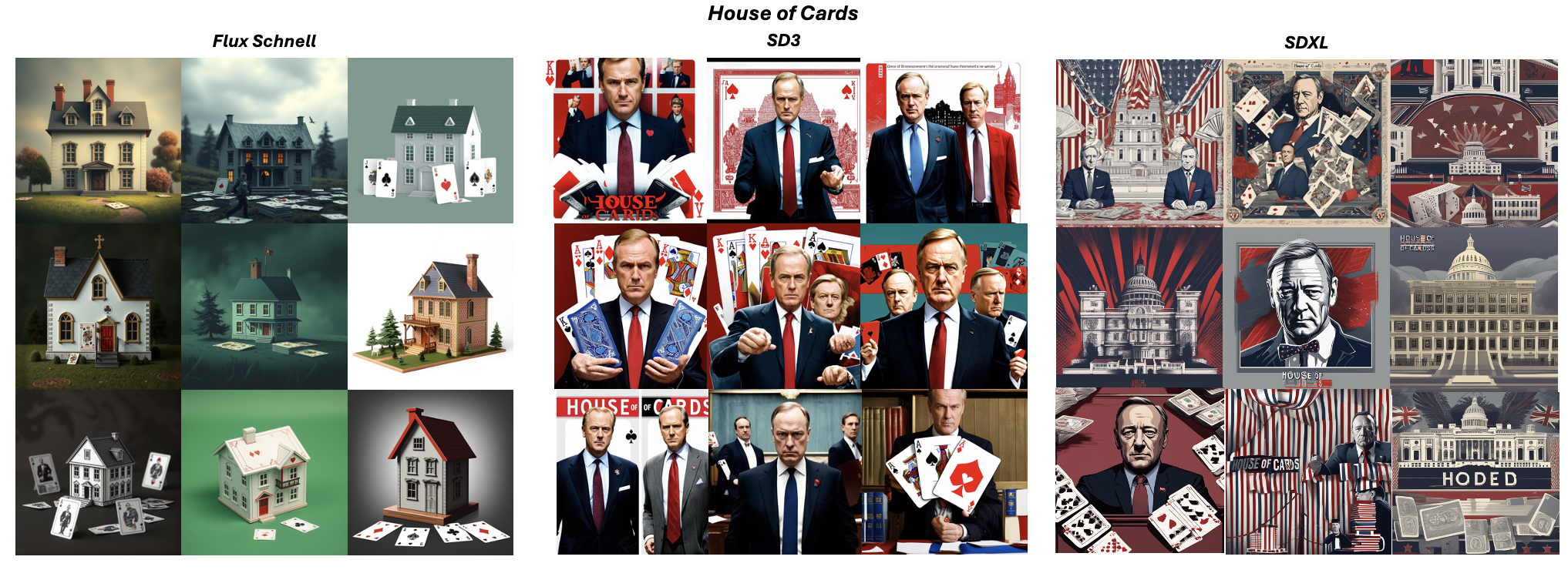}
    \caption{
    \textbf{Prompt:} \emph{House of Cards}. 
    \textbf{Models:} Flux Schnell (left), SD3 (center), SDXL (right). 
    \textbf{CRA:} Flux Schnell = 0.0;\; SD3 = 1.0;\; SDXL = 1.0. 
    \textbf{CRT:} Flux Schnell = 0.00;\; SD3 = 0.49;\; SDXL = 0.95.
    }
    \label{fig:house_of_cards}
\end{figure*}

\begin{figure*}[ht!]
    \centering
    \includegraphics[width=0.8\textwidth]{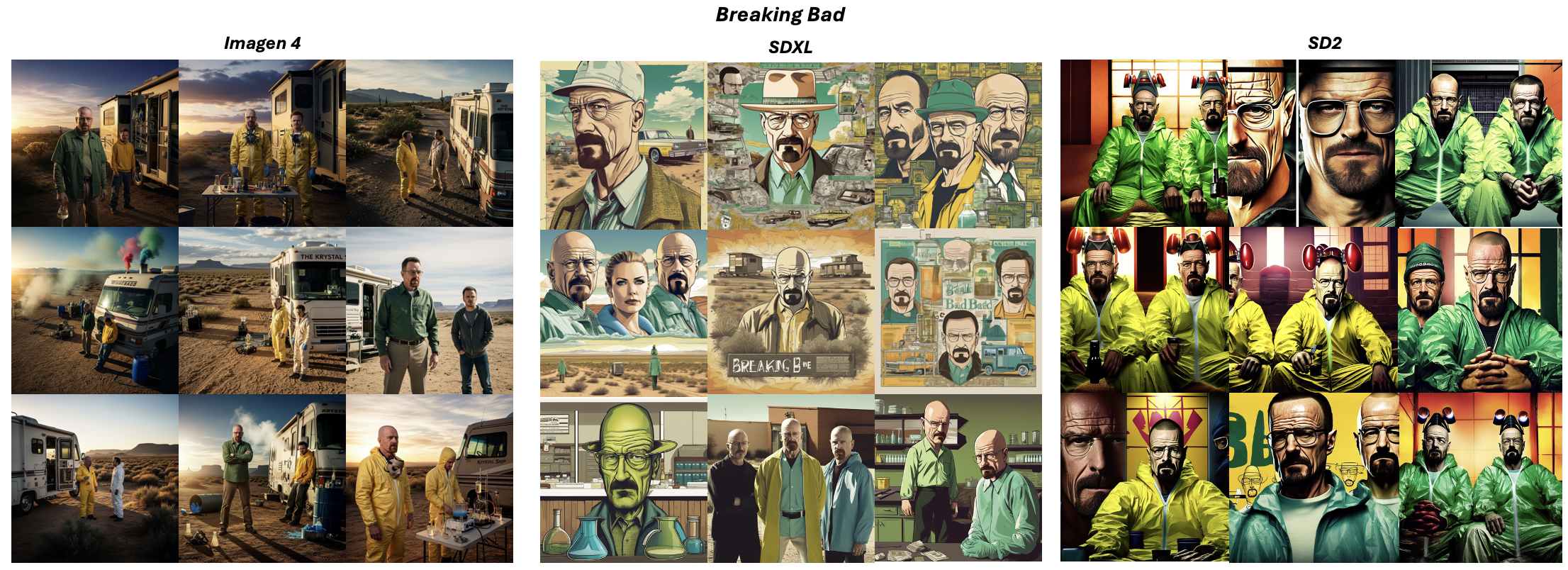}
    \caption{
    \textbf{Prompt:} \emph{Breaking Bad}. 
    \textbf{Models:} Imagen~4 (left), SDXL (center), SD2 (right). 
    \textbf{CRA:} Imagen~4 = 1.0;\; SDXL = 1.0;\; SD2 = 1.0. 
    \textbf{CRT:} Imagen~4 = 0.59;\; SDXL = 0.87;\; SD2 = 0.21.
    }
    \label{fig:breaking_bad}
\end{figure*}

\clearpage
\section{Coverage in Moving-image References.}

\begin{figure}[ht!]
    \centering
    \includegraphics[width=0.55\textwidth]{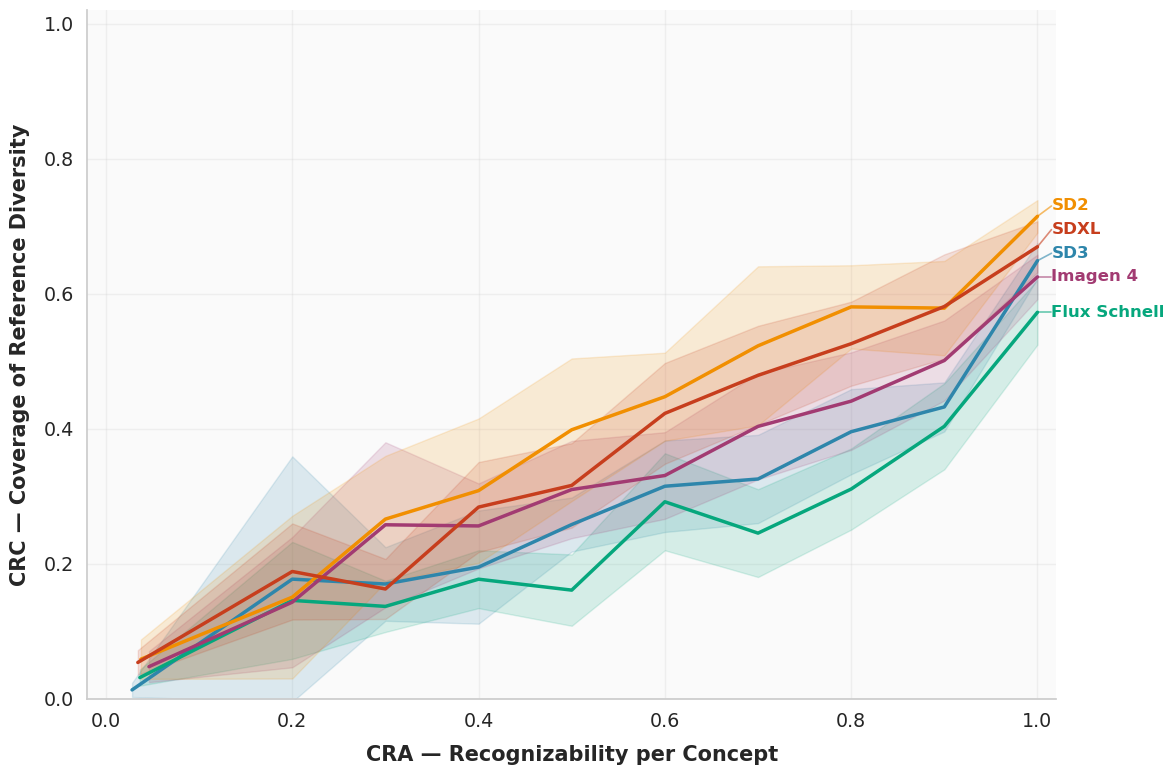}
    \caption{CRA–CRC across models for \textbf{moving-image references}. References are grouped by CRA bins (1.0, 0.9, 0.8, etc.) and plotted against their average CRC values.}
    \label{fig:cra_crc}
\end{figure}

While still-image references are represented by a single canonical image, moving-image references include multiple reference depictions. We therefore further introduce the \textit{Cultural Reference Coverage (CRC} metric, which measures the proportion of these depictions for which at least one generated sample has $s_i > \tau$. Higher CRC indicates broader visual coverage of a reference, while lower values suggest narrower visual diversity. CRC measures how broadly a model reproduces the range of characteristic visual motifs associated with a moving-image cultural reference, indicating whether it generates diverse variants rather than repeatedly depicting the same visual instance. In Fig.~\ref{fig:cra_crc}, we group each recognized cultural reference into its corresponding CRA bin (e.g., 1.0, 0.9, 0.8), aggregating all references recognized 10/10, 9/10, 8/10 times, respectively. We then plot the average CRC of each bin to examine how visual coverage changes with increasing recognition consistency and assess whether models that recognize a concept more reliably also depict it more diversely. CRC increases consistently with CRA across models, but with clear differences in slope that reveal how efficiently each model expands its visual coverage as recognition improves. SD2 maintains the steepest growth curve, indicating that as it learns to recognize more moving-image references, it also diversifies its representations rather than converging on a single visual template. SD3 follows a similar but slightly shallower trend, achieving broad recognition with moderately reduced coverage diversity. Imagen 4 and SDXL show moderate correlation between recognition and coverage, suggesting that while they identify key visual cues, their generative variability remains constrained. Flux Schnell exhibits the lowest overall CRC values, confirming that its outputs, though often visually distinct, cover a narrower range of characteristic reference variants. These results demonstrate that high recognition does not necessarily entail wide coverage: SD2 and SD3 balance these dimensions most effectively, capturing both the identity and diversity of moving-image cultural references.

\section{Perturbation Experiments: Methodological Details}
\label{sec:llm_prompts}

To systematically generate controlled linguistic variations for the \textit{prompt perturbation experiments}, we used two variants of the Llama-3.2 family: the text-only \textbf{Llama-3.2-11B} for synonym substitutions and the \textbf{multimodal Llama-3.2-Vision-11B} for literal-description prompts. Among the models tested (including Llama-3.1-8B, Qwen 2.5 VL, Mistral-7B, and Llama-3.2-Vision-11B), these provided the most coherent and semantically faithful outputs across both perturbation types. All generated prompts were manually screened to ensure that the intended visual referent remained unchanged.

We implemented two prompt-generation methods: (\textit{i}) a \textbf{text setup} for synonym substitutions, and (\textit{ii}) a \textbf{multimodal VQA setup} for literal descriptions. This section provides the exact instructions used for each perturbation type.

\paragraph{(a) Synonym Variant.}
This perturbation replaces one content word (noun, adjective, or verb) from the original title with a close synonym while keeping all other words intact. Llama-3.2-11B was given the following instruction:

\begin{lstlisting}
You are given a text input. Replace only one of the content words
(noun, adjective, or verb) with a single-word synonym that keeps
the rest of the phrase identical. Do not alter word order or add
new terms.

Example:
Input title: "The Scream"
Expected output: "The Shriek"
\end{lstlisting}

\paragraph{(b) Literal Description.}
This perturbation was implemented using a \textbf{multimodal VQA setup}. The multimodal model \textbf{Llama-3.2-Vision-11B} received an image and produced a textual description. For still-image references, we provided the canonical Wikidata reference image; for moving-image references, we identified near-duplicate clusters using \textsc{SSCD} (similarity $\geq 0.90$) and selected a representative from the largest cluster. The model was instructed as follows:

\begin{lstlisting}
You are given an image representing an iconic artwork or scene.
Write a short, objective description of what is visually depicted.
Do not name the artwork, artist, location, or any other identifying
details. Focus only on composition, objects, figures, and perceptual
elements.

Example:
Input image: Image of "The Scream" by Edvard Munch
Expected output: "Painting of a figure standing on a bridge
clutching its face with an open mouth beneath a sky with red and
orange waves."
\end{lstlisting}

\section{Recognition Retention under Prompt Perturbations}
\label{sec:pertrubation_results}

To complement the perturbation analysis presented in \cref{sec:prompt_perturbations}, we report the absolute recognition counts and retention rates for each model under synonym and description substitutions. For each prompt variant, we measure how many cultural references remain recognized relative to the baseline prompts used in the main evaluation. Retention percentages indicate the proportion of references that remain recognizable after lexical modification.

\begin{table}[ht!]
\centering
\small
\caption{Recognition retention under prompt perturbations. Cultural references recognized with the original prompt (\emph{Before}) and retained after synonym or description substitutions. Percentages denote retention relative to the baseline.}
\label{tab:appendix_recognition_retention}

\vspace{2pt}
\textbf{(a) Still Image Cultural References}\\[3pt]
\begin{tabular}{lccc}
\toprule
\textbf{Model} & \textbf{Before} & \textbf{Synonym Retained (\%)} & \textbf{Description Retained (\%)} \\
\midrule
Flux Schnell & 150 & 18 (12.0) & 41 (27.3) \\
\textbf{Imagen 4} & \textbf{233} & \textbf{73 (31.3)} & \textbf{82 (35.2)} \\
SD2  & 183 & 49 (26.8) & 41 (22.4) \\
SD3  & 200 & 23 (11.5) & 25 (12.5) \\
SDXL & 214 & 51 (23.8) & 59 (27.6) \\
\bottomrule
\end{tabular}

\vspace{6pt}

\textbf{(b) Moving Image Cultural References}\\[3pt]
\begin{tabular}{lccc}
\toprule
\textbf{Model} & \textbf{Before} & \textbf{Synonym Retained (\%)} & \textbf{Description Retained (\%)} \\
\midrule
Flux Schnell & 266 & 45 (16.9) & 123 (46.2) \\
\textbf{Imagen 4} & \textbf{320} & \textbf{108 (33.8)} & \textbf{140 (43.8)} \\
SD2  & 340 & 40 (11.8) & 84 (24.7) \\
SD3  & 343 & 70 (20.4) & 137 (39.9) \\
SDXL & 268 & 41 (15.3) & 115 (42.9) \\
\bottomrule
\end{tabular}

\end{table}

Across both still and moving-image references, lexical perturbations substantially reduce recognition relative to the original prompts. Synonym substitutions generally lead to the largest drops in recognition, whereas literal descriptions retain a larger portion of recognizable references. Consistent with the trends discussed in the main text, \textit{Imagen 4} maintains the highest retention rates across both perturbation types, indicating greater robustness to changes in prompt wording.

\clearpage

\section{Examples of Residual Duplicates in LAION}
\label{sec:dedup_examples}
\begin{figure}[ht!]
  \centering
  \includegraphics[width=0.9\linewidth]{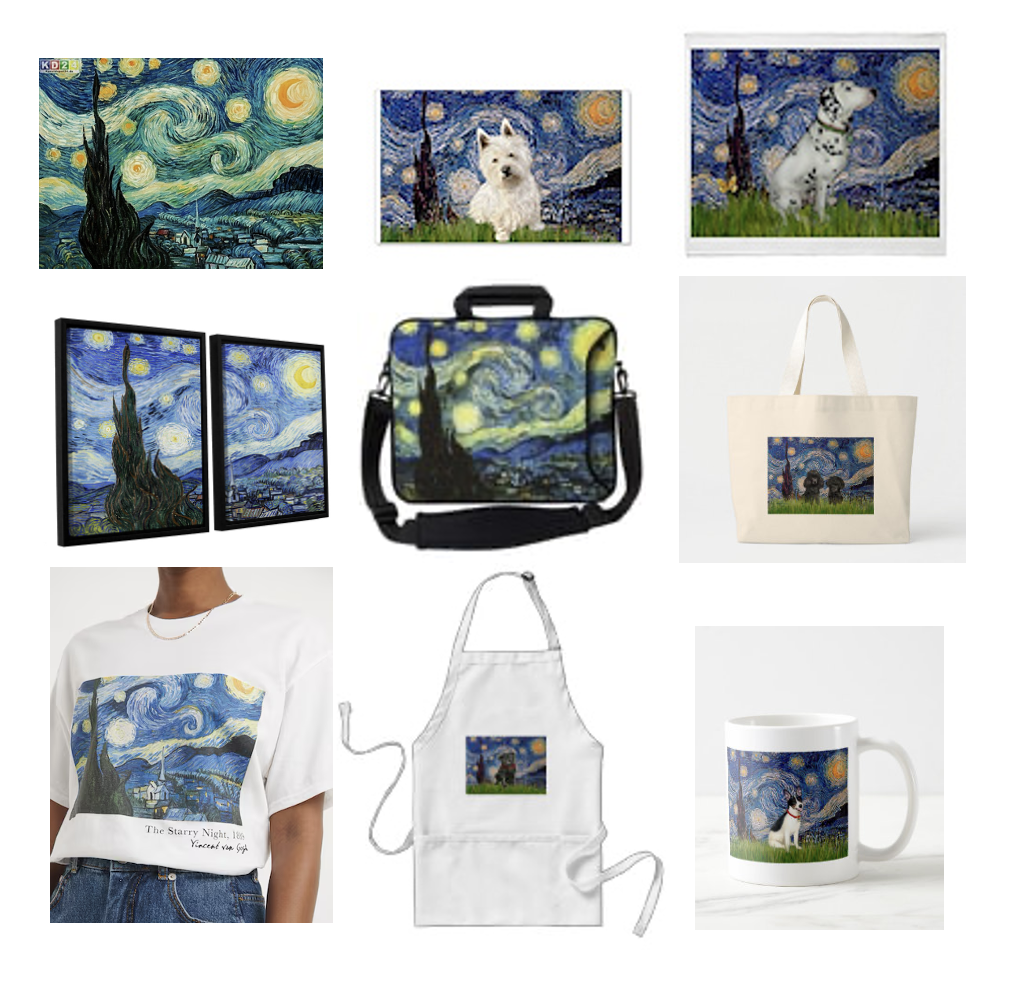}
  \caption{%
    Examples of derivative reproductions of \emph{The Starry Night} found in LAION. 
    Even after near-duplicate removal, visually varied but semantically redundant products (e.g., shirts, mugs, posters) remain.
  }
  \label{fig:dedup_examples}
\end{figure}

\cref{fig:dedup_examples} illustrates the persistence of semantically redundant but visually distinct instances of \emph{The Starry Night} in LAION, even after applying near-duplicate removal. While exact pixel-level copies are filtered out, the dataset still contains numerous derivative reproductions, such as posters, mugs, T-shirts, tote bags, and other products that replicate Van Gogh’s composition in slightly altered visual forms. These examples highlight how cultural artifacts that have entered the domain of mass reproduction generate dense clusters of related imagery in the training corpus. Such residual redundancy amplifies the statistical association between the caption “Starry Night” and specific visual features (e.g., swirling skies, cypress silhouette), thereby reinforcing this link in the model’s latent space despite deduplication.

\section{Correlation Analysis}
\label{sec:correlation_analysis}

As shown in \cref{tab:correlation_analysis}, both still and moving image references exhibit significant correlations between Cultural Reference Alignment (CRA) and several cultural and training-data–related features. For still-image references, \textbf{creation date} ($\rho=-0.63$) and \textbf{text uniqueness} ($\rho=0.50$) emerge as the strongest predictors, indicating that older and linguistically distinctive references are more consistently recognized. \textbf{Image memorability} and training-related factors such as \textbf{number of deduplicated text-image pairs} also show moderate positive correlations. In the moving-image reference setting, \textbf{text uniqueness} remains the dominant factor, while the effects of temporal and visual properties diminish. Overall, the results in \cref{tab:correlation_analysis} confirm that CRA is primarily driven by the distinctiveness and specificity of cultural cues rather than by data volume.

\begin{table}[ht!]
\centering
\scriptsize
\setlength{\tabcolsep}{6pt}
\begin{tabular}{lrrrr}
\toprule
\textbf{Feature} &
\multicolumn{2}{c}{\textbf{Still-images}} &
\multicolumn{2}{c}{\textbf{Moving-images}} \\
\cmidrule(lr){2-3} \cmidrule(lr){4-5}
& $\rho$ & $p$ & $\rho$ & $p$ \\
\midrule
\textbf{Creation Date}           & \textbf{-0.626} & \textbf{0.00} & \textbf{-0.101} & \textbf{0.05} \\
\textbf{Text Uniqueness}         & \textbf{0.496}  & \textbf{0.00} & \textbf{0.444}  & \textbf{0.00} \\
\textbf{Image Memorability}      & \textbf{0.315}  & \textbf{0.00} & 0.071  & 0.17 \\
\textbf{Number of deduplicated text–image pairs} & \textbf{0.158}  & \textbf{0.01} & \textbf{0.192}  & \textbf{0.00} \\
\textbf{Text Concreteness}       & \textbf{0.157}  & \textbf{0.01} & 0.037  & 0.47 \\
\textbf{Popularity}              & -0.120 & 0.08 & \textbf{0.160}  & \textbf{0.00} \\
Word Memorability                & 0.060  & 0.33 & -0.051 & 0.32 \\
Image Uniqueness                 & -0.050 & 0.41 & -0.020 & 0.12 \\
\bottomrule
\end{tabular}
\caption{Spearman correlations between features and Cultural Reference Alignment (CRA) for still and moving image references. Significant results ($p<0.05$) are shown in bold.}
\label{tab:correlation_analysis}
\end{table}

\end{document}